\documentclass{article}

\usepackage[usenames,dvipsnames]{xcolor}
\definecolor{shadecolor}{gray}{0.9}

\usepackage{graphicx}

\usepackage{booktabs}

\usepackage[algoruled, algo2e]{algorithm2e}
\usepackage{algorithm}
\usepackage{algorithmic}

\usepackage[numbers]{natbib}

\usepackage[colorlinks,linktoc=all]{hyperref}
\usepackage[all]{hypcap}
\hypersetup{citecolor=BurntOrange}
\hypersetup{linkcolor=MidnightBlue}
\hypersetup{urlcolor=MidnightBlue}

\usepackage[acronym,smallcaps,nowarn]{glossaries}

\usepackage{nccmath}

\newacronym{mcmc}{mcmc}{Markov chain Monte Carlo}
\newacronym{smc}{smc}{sequential Monte Carlo}

\newacronym{cs-pca}{cs-pca}{clinically significant prostate cancer}
\newacronym{mr}{mr}{magnetic resonance}
\newacronym{emrt}{emrt}{end-to-end Magnetic Resonance Triaging}

\newacronym{kspace-net}{kspace-net}{kspace-net}
\newacronym{auroc}{auroc}{area under the receiver operating characteristic curve}

\newcommand{\kl}[1]{\textsc{KL}\left(#1\right)}

\newcommand{\norm}[1]{\left\lVert#1\right\rVert}

\def\eqref#1{equation~\ref{#1}}

\def\1{\bm{1}}

\def\rvh{{\mathbf{h}}}

\def\rvm{{\mathbf{m}}}

\def\rvs{{\mathbf{s}}}

\def\rvx{{\mathbf{x}}}
\def\rvy{{\mathbf{y}}}
\def\rvz{{\mathbf{z}}}

\DeclareMathAlphabet{\mathsfit}{\encodingdefault}{\sfdefault}{m}{sl}
\SetMathAlphabet{\mathsfit}{bold}{\encodingdefault}{\sfdefault}{bx}{n}

\newcommand{\E}{\mathbb{E}}

\newcommand{\R}{\mathbb{R}}

\DeclareMathOperator*{\argmax}{arg\,max}

\newcommand{\kspace}{\textit{k}-space }

\usepackage[preprint]{neurips_2022}

\usepackage[utf8]{inputenc} 
\usepackage[T1]{fontenc}    
\usepackage{url}            

\usepackage{amsfonts}       
\usepackage{nicefrac}       
\usepackage{microtype}      
\usepackage{xcolor}         
\usepackage[nameinlink]{cleveref}
\usepackage{soul}

\title{On the Feasibility of Machine Learning Augmented 
Magnetic Resonance for Point-of-Care Identification of 
Disease}

\author{%
  Raghav Singhal$^{1}$\thanks{Equal Contribution. $^{1}$ Department of Computer Science, New York University, New York, NY. $^2$ Center for Advanced Imaging Innovation and Research (CAI2R), Department of Radiology, New York University Grossman School of Medicine, New York, NY, United States. $^3$ Center for Data Science, New York University, New York, NY, United States. Correspondence to: Raghav Singhal <rsinghal@nyu.edu>.} \\
  \And
  Mukund Sudarshan$^{1, *}$ \\
  \And
  Anish Mahishi$^{1}$ \\
  \And
  Sri Kaushik$^{1}$ \\
  \And
  Luke Ginnochio$^{2}$ \\ 
  \And 
  Angela Tong$^{2}$ \\
  \And 
  Hersh Chandarana$^{2}$ \\
  \And 
  Daniel Sodickson$^{2}$ \\
  \And 
  Rajesh Ranganath$^{1,3}$ \\
  \And 
  Sumit Chopra$^{1,2}$
}

\begin{document}

\maketitle

\begin{abstract}
  Early detection of many life-threatening diseases (e.g., prostate and breast cancer) within at-risk population can improve clinical outcomes and reduce cost of care.
  While numerous disease-specific ``screening" tests that  are closer to Point-of-Care (\textsc{poc}) are in use for this task, their low specificity results in unnecessary biopsies, leading to avoidable patient trauma and wasteful healthcare spending.
  On the other hand, despite the high accuracy of Magnetic Resonance (\textsc{mr}) imaging in  disease diagnosis, it is not used as a \textsc{poc} disease identification tool because of poor accessibility. 
  The root cause of poor accessibility of \textsc{mr} stems from the requirement to reconstruct high-fidelity images, as it necessitates a lengthy and complex process of acquiring large quantities of high-quality \textit{k}-space measurements. 
  In this study we explore the feasibility of an \textsc{ml}-augmented \textsc{mr} pipeline that directly infers the disease sidestepping the image reconstruction process. 
  We hypothesise that the disease classification task can be solved using a very small tailored subset of \textit{k}-space data, compared to image reconstruction.
  Towards that end, we propose a method that performs two tasks: 1) identifies a subset of the \textit{k}-space that maximizes disease identification accuracy, and 2) infers the disease directly using the identified \textit{k}-space subset, bypassing the image reconstruction step.  
  We validate our hypothesis by measuring the performance of the proposed system across multiple diseases and anatomies. 
  We show that comparable performance to image-based classifiers, trained on images reconstructed with full $k$-space data, can be achieved using small quantities of data: $8\%$ of the data for detecting multiple abnormalities in prostate and brain scans, and $5\%$ of the data for detecting knee abnormalities. 
  To better understand the proposed approach and instigate future research, we provide an extensive analysis and release code. 
  
\end{abstract}

\section{Introduction}
\label{sec:introduction}
Early and accurate identification of  several terminal diseases, such as breast cancer \citep{marmot2013benefits}, prostate cancer \citep{ilic2018prostate}, and colon cancer \citep{winawer1997colorectal}, within the at-risk population followed by appropriate intervention  leads to favorable clinical outcomes for patients by reducing mortality rates \citep{stang2018impact} and reducing cost of care. 
In the current standard-of-care this goal is accomplished by subjecting at-risk but otherwise asymptomatic individuals within the population to clinical tests (a.k.a., ``screening tests'') that identify the presence of the disease under consideration: a process formally referred to as ``Population-Level Screening (\textsc{pls}).'' 
Desiderata for an effective screening test are: 1) it should be safe for use, 2) it should be accurate (have high sensitivity and specificity), and 3) it should be fast and easily accessible to facilitate use at population-level. 
While numerous disease-specific screening tests that are administered closer to point-of-care (\textsc{poc}), and hence are accessible at population-level, have been proposed and are in use, most of them do not satisfy all the three requirements mentioned above. 
For instance, prostate cancer \citep{slatkoff2011psa} and breast cancer \citep{elmore1998ten} have accessible tests, but these tests have low specificity, as shown by multiple clinical trials \cite{kilpelainen2010false,fenton2018prostate}.
Low specificity of these tests results in over-diagnosis and over-treatment of patients leading to many unnecessary, risky, and expensive followup procedures, such as advanced imaging and/or invasive tissue biopsies.
This in-turn causes avoidable patient trauma and significant wasteful healthcare spending \citep{kilpelainen2010false,lafata2004economic, brodersen2013long,taksler2018implications}.

Magnetic Resonance Imaging (\textsc{mri}) has been shown to be a highly effective tool for accurately diagnosing multiple diseases, especially those involving soft-tissues 
\citep{rosenkrantz2012prostate,garcia2015detection,thompsonuro2016,wysockbju2016,rastinehad2014improving,brown2018healthtech}. While traditionally \textsc{mri} is 
used to validate clinical hypotheses under a differential diagnosis regime and is typically used as last in line tool, multiple recent studies have proposed new 
disease specific data acquisition protocols that can potentially
make \textsc{mr} useful for the purpose of early disease identification \citep{eldred2021population,wallis2017role,marino2018multiparametric,biederer2017screening}. These studies have shown that \textsc{mr} can outperform the screening tests being used as part of current standard-of-care. 
However, despite its proven clinical benefits, the challenges associated with the accessibility of \textsc{mri}, limits its widespread use at population-level. 
\ul{As such, there is an unmet need for a \textsc{poc} tool that has the diagnostic accuracy of \textsc{mr} and yet is readily accessible at population-level}. 
Such a tool can have widespread positive impact on the standard-of-care for multiple life threatening diseases. 
Specifically, \textbf{patients} will receive improved care via easy access to \textsc{mr} technology outside of the high-friction specialized environments of imaging centers for early and accurate identification of diseases; \textbf{radiologists} will see an increased diagnostic yield of expensive followup scans since the tool will ensure that only patients with high-likelihood of the disease undergo full diagnostic imaging; and \textbf{health system} will see a reduction in overall cost of care with the decrease in the number of unnecessary expensive follow-up diagnostics and treatment procedures.

\begin{figure}[th]\label{fig:prostate_deterioration}
\centering
\includegraphics[width=0.8\textwidth]{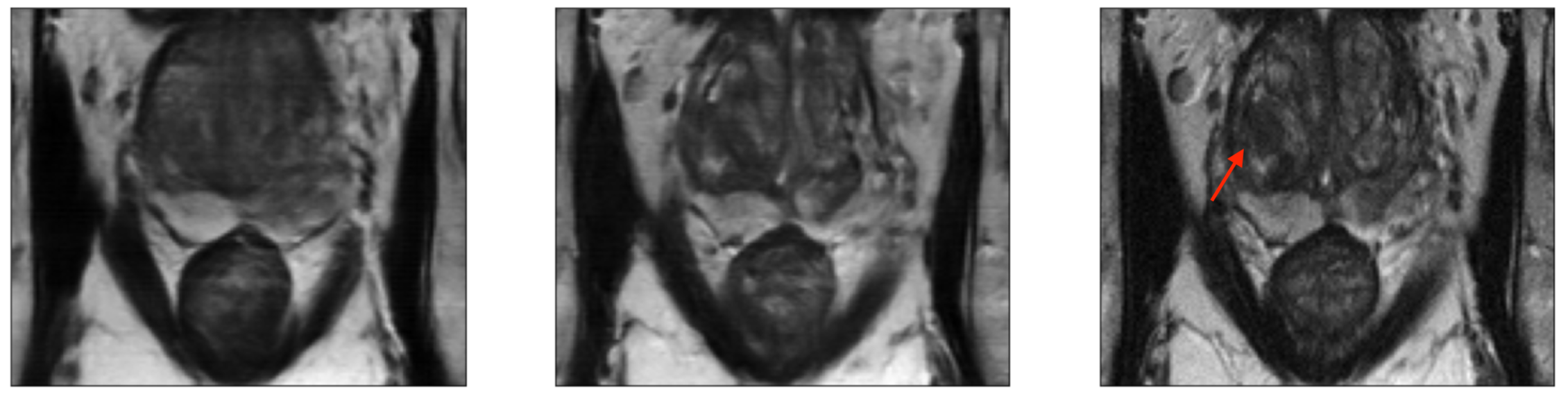}
\caption{\textit{\textbf{Figure showing deterioration in the quality of 
reconstructed images with increasing sampling factors (from left to right)}. 
Lesion visible (red arrow) on image reconstructed from the 
fully-sampled \textit{k}-space data (\textbf{right panel}) is not visible 
in the image reconstructed from $12.5\%$ (\textbf{middle panel}) or $8\%$ sampled
data (\textbf{left panel}) when reconstructed with state-of-the-art reconstruction
 methods. }}
\end{figure}

\begin{figure*}[t]\centering
\includegraphics[width=\columnwidth]{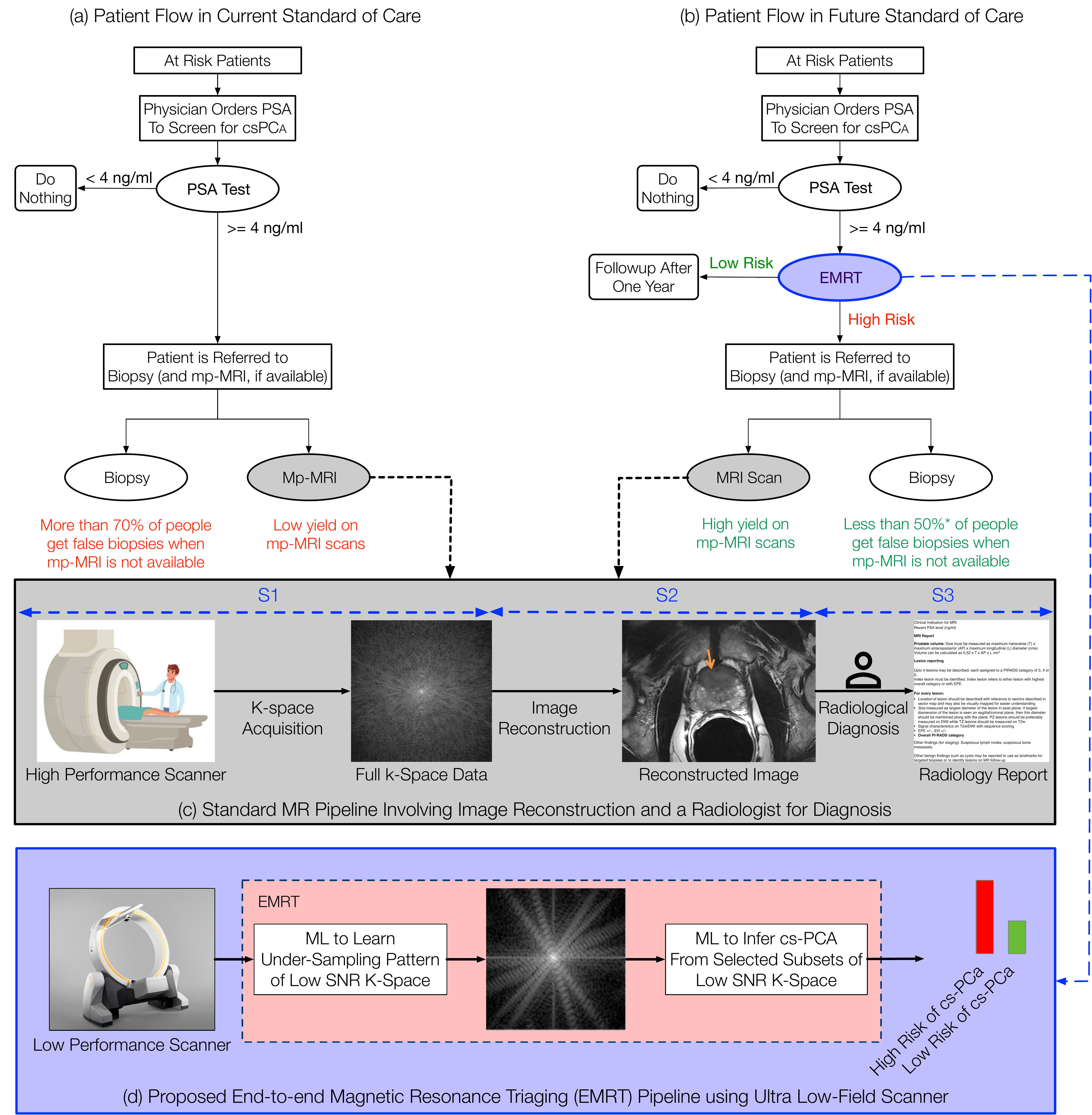}
\caption{\textit{\textbf{Overview of current and proposed standards of care for Prostate Cancer:}
Panel (a) depicts the current practice of testing for clinically significant prostate cancer 
(\textsc{cs-pca}),
 which involves testing at-risk patients using a 
 \textsc{psa} 
 test followed by an expensive multi-parametric \textsc{mri} (Panel (c)) and a biopsy. In Panel (b), with our proposed 
 triaging tool, patients who have a high \textsc{psa} score undergo a subsequent test with 
 the \textsc{emrt} embedded ultra-low field \textsc{mr} device (Panel (d)). With the use of 
 the triaging device, only high risk patients get the expensive and inaccessible 
 multi-parametric \textsc{mri} and invasive biopsy, reducing waste in the healthcare system 
 and preventing as many as 38\% of the biopsies
  \citep{slomski2017avoiding}.}}
\label{fig:mr_pipeline}
\end{figure*}

To understand the reason behind poor accessibility of \textsc{mr}, we first shed light on the workings of the pipeline. 
Figure \ref{fig:mr_pipeline} (c) depicts the full \textsc{mr} pipeline. 
\textsc{mr} imaging is an indirect imaging process in which the \textsc{mr} scanner subjects the human body with magnetic field and radio-frequency signals and measures the subsequent electromagnetic response activity from within the body. 
These measurements are collected in the Fourier space, also known as $k$-space (see section $5.4$ in \cite{brown2011mri}) (stage {\bf S1} in figure \ref{fig:mr_pipeline} (c)). 
The $3$D volumetric image of the anatomy is reconstructed from these \textit{k}-space measurements using a multi-dimensional inverse Fourier transform (stage {\bf S2}). 
The images are then finally interpreted by sub-specialized radiologists who render the final diagnosis (stage {\bf S3}). 
The reason behind \textsc{mr}'s diagnostic success is its ability to generate these high-fidelity images with excellent soft-tissue contrast properties, because such images enable the human radiologists to easily discern the pathology accurately. 
The quality of the images is directly related to the quantity and the quality of the \textit{k}-space measurements acquired: large quantities of high-quality measurements results in a high-quality image.
This in-turn necessitates the need for 1) expensive specialized scanners installed in special purpose imaging centers to collect large quantities of high-quality \textit{k}-space data, 2) execution of long and complex data acquisition protocols to reconstruct the high-fidelity images exhibiting multiple contrasts, and 3) sub-specialized radiologists to interpret the reconstructed images. 
All these factors prevent \textsc{mr} scanning to be used as a tool closer to \textsc{poc} for early and accurate disease identification. 
Instead its use is predominantly limited to validating a clinical hypothesis at the end of the diagnostic chain. 
With the motivation of improving accessibility of \textsc{mr}, researchers have proposed multiple solutions to simplify the pipeline. 
These include designing novel acquisition protocols to acquire the \textit{k}-space data \citep{kasivisvanathan2018mri,eldredjamaonco2021},
learning the under-sampling pattern over \textit{k}-space data matrices so that the image quality is not compromised \cite{bahadir2019learning,zhang2019reducing,weiss2020joint,huijben2019deep}, 
faster data acquisition and image reconstruction from the under-sampled \textit{k}-space data, and for simultaneous classification and image-reconstruction using under-sampled \textit{k}-space data
\cite{johnson2022deep, 
lustig2007sparse,lustig2008compressed,zbontar2018fastmri,muckley2021results,hammernik2018learning}.
While these efforts have expedited the data acquisition process, the requirement to generate high-fidelity images still necessitates the use of expensive scanners and the need for a sub-specialized radiologist to interpret them. 
Furthermore, image generation also imposes limits on how much one can under-sample the $k$-space. 
For instance, \citep{muckley2021results} reports that reconstructed images started missing clinically relevant pathologies if we sample less than $25\%$ of the data. 
This phenomenon can be observed in Figure \ref{fig:prostate_deterioration}, which shows images reconstructed by a state-of-the-art reconstruction model \cite{sriram2020end} using different levels of sampling. 
A clearly visible lesion in the high resolution image is barely visible in the image generated using $8\%$ data. 

This work is motivated by the goal of making the benefits of \textsc{mr} diagnostics available for population-wide identification of disease.
Towards that end, we ask the following questions: 
``If the clinical goal is to merely identify the presence or absence of a specific disease (a binary task accomplished by a typical screening test), is it necessary to generate a high-fidelity image of the entire underlying anatomy? 
Instead, can we build an \textsc{ml} model that can accurately provide the final answer (whether a disease is present or not) from a carefully selected subset of the \textit{k}-space data?"
\ul{Specifically, we hypothesize that when the task is to infer the presence of a disease (a binary decision), we do not need all the \textit{k}-space measurements that are otherwise acquired to generate a high-fidelity image. Instead, we can train an \textsc{ml} system that can accurately provide the binary answer directly from a carefully tailored small fraction of degraded \textit{k}-space data that can potentially be acquired using low-grade inexpensive scanning devices.}
To validate the above hypothesis, one needs to answer the following key questions:

\begin{itemize}
    \item [{\bf Q1.}]
    Can we build a \textsc{ml} system that can accurately infer the presence of a disease using data from standard \textsc{mr} sequences without generating images?
    
    \item [{\bf Q2.}]
    Can we build a \textsc{ml} system that uses only a small fraction of carefully tailored subset of the \textit{k}-space data to infer the presence of a disease without images? If so, how little data do we need without compromising     performance?
        
    \item [{\bf Q3.}]
    Can we build a \textsc{ml} system that can accurately infer the presence of a disease using degraded \textit{k}-space data without generating images? What are the limits on signal quality we can afford to work with, without compromising performance?
\end{itemize}

Answers to these questions will shed light on the feasibility of making \textsc{mr} scanning accessible outside of its current specialized environments to be potentially used for the purpose of early, efficient, and accurate identification of disease at population-level. 
In this study we answer {\bf Q1}, and {\bf Q2} and leave the answers to {\bf Q3} as future work. 
Towards that end, we first propose a novel deep learning (\textsc{dl}) model that takes as input the raw \textit{k}-space data and generates the final (binary) answer, skipping the image reconstruction step (Section \ref{sec:classifier}). 
We show that it is indeed possible to train a \textsc{ml} model that can directly generate an answer from the \textit{k}-space data without generating an image. 
This result is not surprising because mapping the \textit{k}-space data to image space is accomplished by simply applying an Inverse Fourier Transform (\textsc{ift}) operation on the \textit{k}-space data, which is a deterministic lossless mapping. 
Next, to answer question Q2, we propose a novel \textsc{ml} methodology that can accurately infer the presence of a disease directly from a small tailored subset of the \textit{k}-space data, side-stepping the image reconstruction step (Section \ref{sec:emrt}). 
We call this methodology the {\bf E}nd-to-end {\bf M}agnetic {\bf R}esonance {\bf T}riaging (\textsc{emrt}). 
Figure \ref{fig:mr_pipeline}(d) provides an outline of our methodology in comparison to the current image reconstruction-based pipeline (Figure \ref{fig:mr_pipeline}(c)). 
\textsc{emrt} simultaneously accomplishes two tasks: 
\begin{enumerate}
\item Identifies a small subset of the \textit{k}-space that can provide sufficient signal for accurate prediction of the disease by an \textsc{ml} model, ignoring the quality of the image it will generate. 

\item It then infers the presence of the disease directly using data from only the identified subset of the \textit{k}-space, without generating an image. 
\end{enumerate}

We validate the efficacy of \textsc{emrt} in identifying multiple diseases using scans from multiple anatomies, namely, to detect presence of \textsc{acl} sprains and meniscal tears in slice-level knee \textsc{mr} scans, to detect enlarged ventricles and mass in slice-level brain \textsc{mr} scans, and detect the presence of clinically significant prostate cancer (\textsc{cs-pca}) in slice-level abdominal \textsc{mr} scans.
The knee and brain scans are made available in the Fast\textsc{mri} data set \citep{zbontar2018fastmri} with labels provided by the Fast\textsc{mri}$+$ data set \citep{zhao2021fastmri+}. 
We use an internal data set for the prostate scans acquired as part of clinical exams of real patients at NYU Langone Health system. 
We compare the performance of \textsc{emrt} against two types of benchmark methods. 

Our first benchmark consists of a classifier trained with images reconstructed from fully-sampled \textit{k}-space data. 
Since the prediction accuracy of this benchmark is the best one can hope for from any image-based classifier, we use this comparison to establish the limits of how much one can under-sample the \textit{k}-space and still accurately infer the disease, when not reconstructing images. 
Our results show that \textsc{emrt} can achieve the same level of accuracy as this benchmark using only $5\%$ of the data for knee scans and $8\%$ of the data for brain and prostate scans. 
Our second benchmark is another image-based classifier that uses as input the images reconstructed from an under-sampled \textit{k}-space data using the state-of-the-art image reconstruction models proposed in the literature \citep{sriram2020end,muckley2021results}. 
The motivation behind this experiment was to show that for the same disease identification accuracy, if we by-pass the image reconstruction step, we require a significantly smaller fraction of the \textit{k}-space data in comparison to when we reconstruct images. 
Our results also show that for all under-sampling rates in our experiments, \textsc{emrt} outperforms under-sampled image-reconstruction based benchmarks even though the images are reconstructed using the state-of-the-art reconstruction models.  
Lastly, we also provide an extensive analysis that shed light on understanding the workings of \textsc{emrt}. 
Our contributions include:
\begin{itemize}
    \item \textsc{emrt}: a first-of-its-kind machine learning methodology that identifies a subset of \textit{k}-space that maximizes disease classification accuracy, and then infers the presence of a disease directly from the \textit{k}-space data of the identified subset, without reconstructing images.
    
    \item Rigorous comparison to the state-of-the-art image reconstruction-based benchmark models to prove the efficacy of the proposed methodology. 
    
    \item Extensive analysis of \textsc{emrt} to understand the reasons behind its superior performance. 
    
    \item Release of the code and data used to build \textsc{emrt} with the goal of facilitating further research in end-to-end methods like \textsc{emrt}, that have the potential to transform healthcare. 
\end{itemize}

\section{Clinical Vision}
\label{sec:clinical_vision}
This study is motivated by the overarching goal of making \textsc{mr} scanning accessible outside of its current specialized environments so that its diagnostic benefits can be realized at population-level for early, efficient, and accurate identification of life-threatening diseases. 
We argue that this poor accessibility is rooted in the requirement to generate high-fidelity images, because image generation necessitates the need to acquire large quantities of high-quality \textit{k}-space data (forcing the use of expensive scanners installed in specialized environments running complex data acquisition protocols) and the need for sub-specialized radiologists for interpretation. 
As such we ask a sequence of questions pertaining to \textit{k}-space data requirements for accurate disease identification under the setting when we are not generating intermediate high-fidelity images. 
Answers to questions posed in this study will shed light on the feasibility of whether our end goals can be accomplished. 

Assuming answers to all the questions are favorable, one can imagine an ultra-low-field inexpensive scanning device that is only capable of acquiring small quantities of low quality \textit{k}-space data, from which it is difficult to reconstruct an image that has a clearly discernible pathology. 
However an \textsc{ml} model (embedded within the device) could accurately infer the presence of the disease directly from this data. 
Such an inexpensive system could be used clinically as a triaging tool in the following way: The system is placed in a primary care clinic where it is used to test patients who are known to be at risk of the disease. 
Patients for whom the system provides a ``yes'' answer (possibly with some confidence score) are routed for the more thorough followup diagnostic procedures (full \textsc{mr} scan and/or biopsy). 
Others are sent back into the surveillance pipeline for subsequent periodic screening. 
More specifically, in Figure \ref{fig:mr_pipeline}(a) and (b), we depict the utility of such a device when screening for clinically significant prostate cancer (\textsc{cspca}), the second most common reason behind male mortality within the United States.
Figure \ref{fig:mr_pipeline}(a) depicts the current standard of care for \textsc{cspca} screening, where at-risk people are ordered to take the \textsc{psa} test to screen for disease, followed by either an invasive biopsy or a $40$ minute long multi-parametric \textsc{mri} exam (depending on the \textsc{psa} value). 
Unfortunately, the high false positive rate of the \textsc{psa} test, causes unnecessary patient trauma and wasteful healthcare spending as $70\%$ of patients who have a positive \textsc{psa} test can get a negative biopsy. 
In Figure \ref{fig:mr_pipeline}(b), we highlight how the proposed triaging tool can be placed in the pipeline. 
The \textsc{psa} test can be followed up by another test using the ultra-low-field \textsc{emrt} embedded \textsc{mri} device. 
Unlike a full \textsc{mri} exam, the \textsc{emrt}-embedded device will not have to produce an image, just a risk score. 
Such a triaging device can filter high and low risk patients further, and only select the high risk patients for subsequent diagnostics tests such as, full \textsc{mri} and/or biopsy. 
This in-turn will reduce waste in the healthcare system and prevent patient trauma. 

This scenario is not far from the realm of reality, as many organizations are manufacturing such ultra low-field specialized scanners, such as Promaxo \cite{nasri2021office} for the prostate and Hyperfine \cite{hamilton2020hyperfine} for the brain, both of which are approved by the FDA. 
\ul{We note that while we are exploring the feasibility of \textsc{ml} enabled \textsc{mr} scanning that generates an answer without images, the use-case of such a device does not replace the current practice of radiology, which requires the generation of high-fidelity images interpreted by sub-specialized radiologists to render the final diagnosis.} 
Such an imaging and subsequent interpretation exercise is important to render the final diagnosis, for staging and planning treatment \citep{caglic2019multiparametric}. Instead, existence of such a device has the ability to generate alternate use-cases for  \textsc{mr} scanning technology.

\section{Related Work}
\label{sec:related_work}
Applications of deep learning (\textsc{dl}) within \textsc{mr} can be grouped into two categories, namely \textit{image analysis} and \textit{image reconstruction}. 
Under the \textit{image analysis} category, \textsc{dl} models take the spatially resolved gray scale $2$D or $3$D \textsc{mr} images as input and perform tasks like tissue/organ segmentation \citep{akkus2017deep,comelli2021deep} or disease identification \citep{shen2019deep, yoo2019prostate, wong2016artificial, zhang2020deep,padhani2019detecting}. \textsc{dl} models have achieved radiologist level performance in identifying numerous diseases \citep{hirsch2021radiologist,hirsch2020deep,hendrix2022musculoskeletal} and are increasingly being deployed as part of computer aided diagnostic systems \citep{sathiyamoorthi2021deep,winkel2021novel}. For instance, the authors in \citep{labus2022concurrent} examined the effect of \textsc{dl} assistance on both experienced and less-experienced radiologists. The \textsc{dl} assisted radiologist surpassed the performance of both the individual radiologist and the \textsc{dl} system alone. 
These approaches have improved diagnostic accuracy but have so far required high-resolution images that are expensive to produce.

Most methods within the \textit{image reconstruction} category are motivated with the goal of improving the accessibility of \textsc{mr} scanning by reducing the scanning time. 
Towards that end, researchers have proposed a variety of solutions to simplify and expedite the data acquisition process. 
Specifically, researchers have proposed machine learning models to enable rapid reconstruction of spatially resolved 2D images from under-sampled \textit{k}-space data acquired by the scanner \citep{lustig2008compressed,sriram2020end,comelli2021deep}. 
This task requires addressing two key questions, namely, 1) what sampling pattern to choose? and 2) given a sampling pattern, what reconstruction method to choose? 
For the first question, researchers have proposed \textsc{ml} methods that learn the sampling pattern over the \textit{k}-space data matrices so that the image quality is not compromised \citep{bahadir2019learning, zhang2020extending, weiss2020joint, bakker2020experimental, huijben2019deep}.
In another line of work, researchers model the \textit{k}-space acquisition process as a sequential decision making process, where each sample is collected to improve reconstruction performance and used reinforcement learning models to solve the task \citep{pineda2020active,jin2019self,bakker2020experimental}. 
To answer the second question, \textsc{dl} models have been proposed that use under-sampled \textit{k}-space data to reconstruct images of provable diagnostic quality
\citep{lustig2007sparse,lustig2008compressed,muckley2021results,hammernik2018learning,sriram2020end,knoll2020advancing,zhu2018image,recht2020using,lee2017deep,hyun2018deep,cole2020unsupervised}.
Researchers have also proposed \textsc{non-ml}-based solutions to expedite the scanning time for \textsc{mr}. 
These solutions involve the design and execution of novel data acquisition protocols and sequences that enable rapid acquisition of 
the \textit{k}-space data \citep{kasivisvanathan2018mri,eldredjamaonco2021}.
Lastly, to facilitate research in image reconstruction, several data sets and challenges have been released, such as the \textsc{fastmri} \citep{zbontar2018fastmri}, \textsc{fastmri}$+$ \citep{zhao2021fastmri+} and Stanford knee \textsc{mri} with multi-task evaluation (\textsc{skm-tea}) \citep{desai2021skm}.
These data sets provide raw \textit{k}-space measurements for \textsc{mr} scans along with labels of abnormalities associated with those scans.

While these efforts have simplified and expedited the data acquisition process, the requirement to generate high-fidelity images still necessitates the use of expensive scanners and the need for a sub-specialized radiologist to interpret them. 
Furthermore, image generation imposes limits on how much one can under-sample the $k$-space. 

Our work instead studies the problem of using \textsc{dl} models to infer the presence/absence of a disease directly from a small learned subset of the \textit{k}-space data has never been considered.

\section{MR Background and Notation} 
\label{sec:background}
\textsc{mr} imaging is an indirect process, whereby which spatially resolved images of a human subject's anatomy are reconstructed from the frequency space (a.k.a., \textit{k}-space) measurements of the electromagnetic activity inside the subject's body after it is subjected to magnetic field and radio-frequency pulses. 
These measurements are captured by an instrument called a \emph{receiver coil} which is kept in the vicinity of the part of the body whose image is sought. 
The \textit{k}-space measurements from a single-coil are represented as a $2$-dimensional complex valued matrix $\rvx \in \mathbb{C}^{r \times c}$, where $r$ is the number of rows and $c$ is the number of columns.
The spatial image $\rvz$ is reconstructed from the \textit{k}-space matrix by a multi-dimensional  inverse Fourier transform, $\rvz = \mathcal{F}^{-1}(\rvx)$.
We denote by $\rvy \in \{1, \dots, K\}$ the clinically relevant response. In our case $\rvy$ will be a binary response variable ($\rvy \in \{1, 0\}$) indicating the presence/absence of the disease being inferred.

\vspace{-3mm}
\paragraph{Multi-Coil Data:}
In practice, to speed up the data acquisition process, most modern \textsc{mr} scanners acquire measurements in parallel using multiple receiver coils. 
In case of multi-coil acquisition, the \textit{k}-space matrix $\rvx^{mc}$ is $3$-dimensional: $\rvx^{mc} \in \mathbb{C}^{d_c \times r \times c}$ \citep{zbontar2018fastmri}, where $d_c$ is the number of coils used.
The image produced by each coil has a slightly different view of the anatomy, since each coil has different sensitivity to signals arising from different spatial locations.
Multiple methods have been proposed to combine/use these images in ways that are conducive for ingestion into any downstream \textsc{ml} model. 
For instance, a commonly used method that combines these images from different coils into a single aggregate image is called the root-sum-of-squares (\textsc{rss}) method  \cite{larsson2003snr}. 
Given the multi-coil \textit{k}-space matrix, the \textsc{rss} method requires computing the inverse Fourier transform of each coil's \textit{k}-space matrix $\tilde{\rvm}_j = \mathcal{F}^{-1}(\rvx_j)$, and then generating the \textsc{rss} image by 
\begin{align*}
    \tilde{\rvm} = \sqrt{\sum_{j=1}^{N_c} \mid \tilde{\rvm}_j  \mid^{2}}.
\end{align*}
Instead of combining the data from multiple coils in the image space, one can also combine the data in the original \textit{k}-space. 
A methods called Emulated Single Coil (\textsc{esc}) \cite{tygert2020simulating}, aggregates directly the \textit{k}-space data from multiple coils and emulates it to be coming from a single coil. 
This process reduces the dimension of the full matrix $\rvx^{mc} \in \mathbb{C}^{d_c \times r \times c}$ to a matrix $\tilde{\rvx}_{mc} \in \mathbb{C}^{r \times c}$.
In the subsequent discussion pertaining to the direct \textit{k}-space model, we will assume that we are working with the emulated single coil data matrix $\tilde{\rvx}^{mc}$ of dimensions $r \times c$.

\vspace{-3mm}

\begin{figure}[!h]
\small
\centering
\vspace{-1.5in}
  \includegraphics[width=0.8\textwidth]{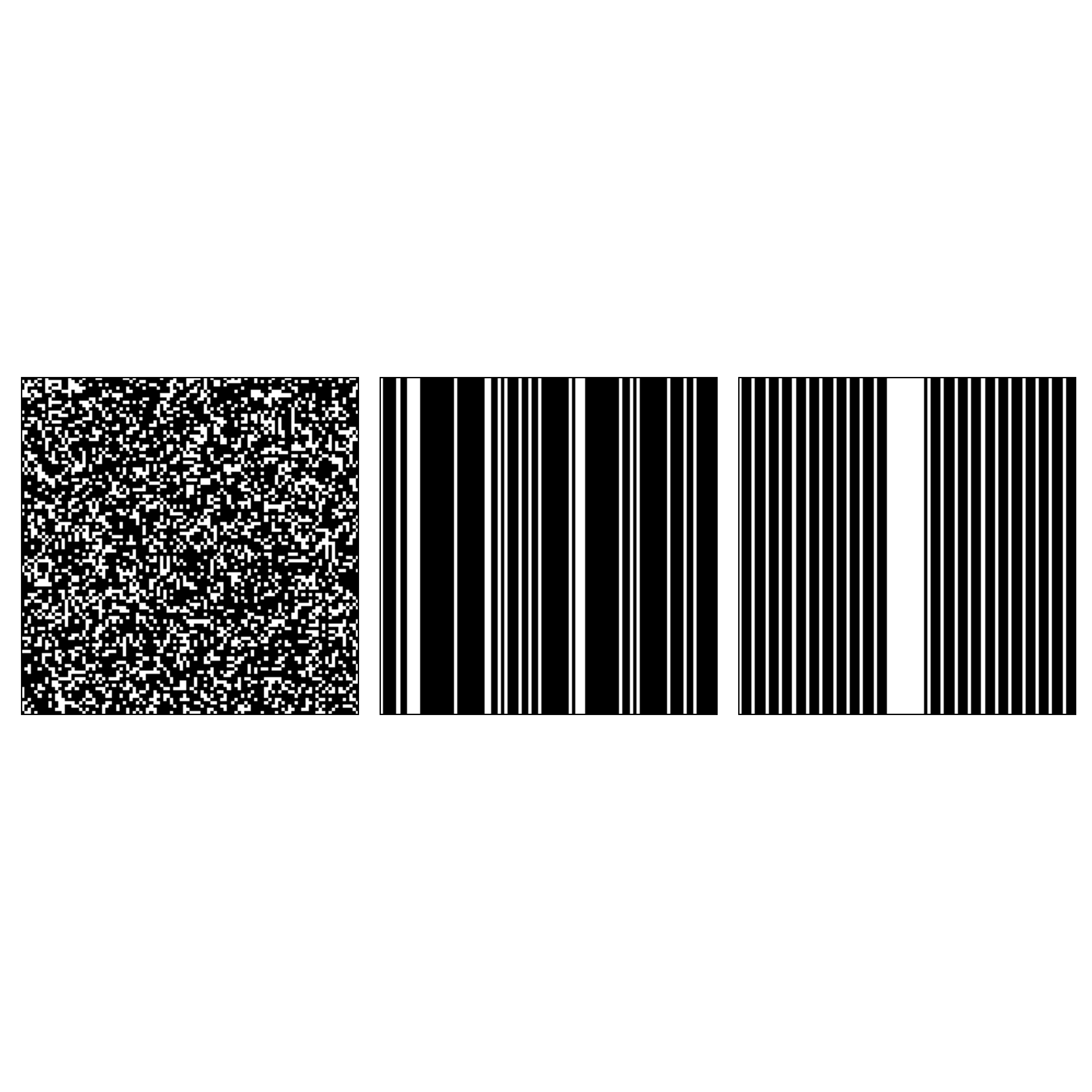}
\vspace{-1.5in}  
\caption{\textit{{\bf Examples of $k$-space sampling patterns:}
The left panel shows a unconstrained sampling pattern with $30\%$ sampling rate, the middle panel shows a random Cartesian sampling pattern with a $30\%$ sampling rate, and the right panel displays an equispaced Cartesian sampling pattern with a $25\%$ sampling rate with sampling. }\label{fig:undersampling patterns}}
\end{figure}

\paragraph{Under-Sampled Data:}
The notion of ``under-sampling'' refers to measuring only a subset of entries in the \textit{k}-space matrix $\rvx$.
We represent the \emph{sampling pattern} using a binary mask matrix $\rvs \in {\{0, 1\}}^{r \times c}$ (sometimes also referred to as \emph{sampling mask}), where $\rvs_{ij} = 1$ if and only if the measurement $\rvx_{ij}$ was acquired.
The under-sampled \textit{k}-space matrix is represented as $\rvx_{\rvs} = \rvx \circ \rvs$, where $\circ$ is element-wise multiplication between the two matrices.
In this work, we constrain the sampling pattern to ``Cartesian'', which consists of sampling the lines of the \textit{k}-space matrix. 
More specifically, for a Cartesian sampling pattern all the elements of some lines of the matrix  $\rvs$ are $0$ and all the elements of other lines are set to $1$. 

See Figure \ref{fig:undersampling patterns} for structure of various sampling patterns. The $k$-space matrix has its origin in the center of the matrix. 
The sampling rate $\alpha$ is defined as the total percentage of measurements.

\subsection{Image-Based Disease Identification using
Deep Learning Models}
\label{sec:image_screen}
The conventional way of using \textsc{dl} models to infer presence of a disease within the \textsc{mr} pipeline involves two steps. 
In the first step, a high-fidelity image is reconstructed using the acquired $k$-space measurements from multiple coils using the \textsc{rss} method, as described above. 
In the second step, the reconstructed image is provided as an input to a \textsc{dl} model that is trained to infer the presence/absence of the disease. 
We refer to this model as \textsc{Model$_{\textsc{rss}}$}. 
This is the best one can hope to achieve when using images and we benchmark the accuracy of \textsc{emrt} against it. 

Since the high-fidelity images used by methods such as \textsc{Model$_{\textsc{rss}}$} requires acquisition of large quantities of high quality \textit{k}-space data, researchers have also proposed to train image-based \textsc{dl} classifiers using images reconstructed from the under-sampled \textit{k}-space data. 
This approach requires one to make decisions at two levels, namely 1) choosing the sampling pattern over the \textit{k}-space, the data from which will be used to reconstruct the image and 2) given the sampling pattern, choosing a method to reconstruct the image. 
Multiple methods have been proposed to learn the sampling pattern \cite{bahadir2019learning,zhang2020extending,weiss2020joint}, and to reconstruct images using  the under-sampled \textit{k}-space data \citep{knoll2020advancing, muckley2021results, hammernik2018learning, sriram2020end, zhu2018image, lee2017deep,hyun2018deep, cole2020unsupervised}. 
We denote the class of these models by \textsc{Model$_{<\textsc{samp}>:<\textsc{recon}>}$}, where <\textsc{samp}> refers to the method used to choose the sampling pattern and <\textsc{recon}> refers to the image reconstruction method. 
We compare the performance of \textsc{emrt} against a variety of these models with different combinations of sampling and reconstruction regimes (Section \ref{sec:experiments}). 

\section{Direct \textit{k}-Space Classifier} 
\label{sec:classifier}
\begin{figure*}[t]
\centering
    \includegraphics[width=\columnwidth]{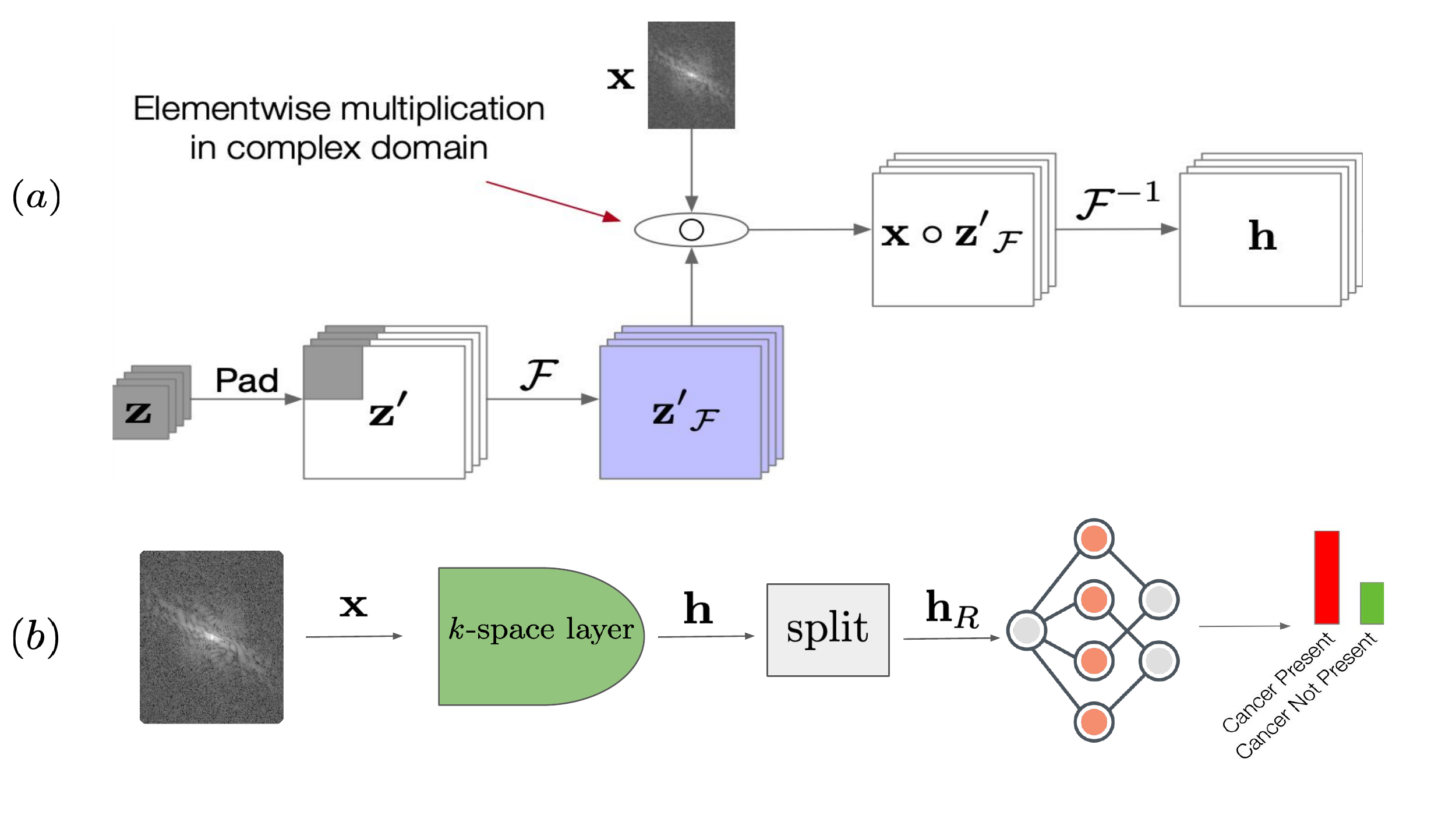}
    \caption{\textit{\textbf{(a) $k$-space layer}: the $k$-space layer
    makes use of the convolution theorem to perform an initial convolution operation between the complex valued \textit{k}-space input $\rvx$ and the kernel $\rvz$. 
    The resulting output is passed through an inverse Fourier transform operation
     to generate real valued feature maps $\rvh_R$ of size $k \times r \times c \times 2$. These 
     feature maps are passed as input to the subsequent layers of \textsc{kspace-net}.
    \textbf{(b) \textsc{kspace-net}}: The \textsc{kspace-net} takes the $k$-space as input 
    followed by the $k$-space layer, then it applies a convolutional architecture on the 
    feature maps $\rvh$ to make a classification.
    }}
    \label{fig:model_arch} 
    \vspace{-4mm}
\end{figure*}

We now describe the proposed \textsc{dl} model that takes as input the \textit{k}-space data and directly generates the final answer without reconstructing an intermediate image. 
The foundational block of our architecture is the convolution theorem, which states that for any given functions $f, g$ we have: 
\begin{equation}
  \mathcal{F}(f * g) = \mathcal{F}(f) \circ \mathcal{F}(g),
  \label{eq:conv_theorem}
\end{equation}
where $\mathcal{F}$ is the Fourier transform, $*$ denotes the convolution operation, and $\circ$ denotes element-wise multiplication.
Multiple researchers in the past have used this operator duality to accelerate convolutions in Convolutional Neural Networks (\textsc{cnn}) \citep{rippel2015spectral,mathieu2013fast}.

Since the \textit{k}-space data is in the frequency domain, we can use Eq. \ref{eq:conv_theorem} to adapt any convolutional neural network architecture to directly use the \textit{k}-space data as input.
Specifically, let $\rvx \in \mathbb{C}^{r \times c}$ denote the complex-valued  $k$-space matrix of size $r \times c$, and
let $\rvz \in \mathbb{C}^{k \times k}$ be the kernel with which we want to convolve the input $\rvx$.
We accomplish this convolution by first zero-padding the kernel to the right and bottom to create a kernel $\rvz' \in \mathbb{C}^{r \times c}$ which is of the same size as the input (see Figure \ref{fig:model_arch}).
We then take the Fourier transform of the padded kernel $\rvz'$, such that  $\rvz'_{\mathcal{F}} = \mathcal{F}(\rvz')$ is in the frequency space. 
Using Equation \ref{eq:conv_theorem}, we compute the convolution between the input $\rvx$ and the kernel $\rvz$ by taking the inverse Fourier transform of the element-wise multiplication of  $\rvx$ and $\rvz'_{\mathcal{F}}$: 
\begin{equation}
    \rvh = \mathcal{F}^{-1}(\rvx) * \rvz = \mathcal{F}^{-1} (\rvx \circ \rvz'_{\mathcal{F}}).
\end{equation}
The matrix $\rvh \in \mathbb{C}^{r \times c}$ is a complex matrix in the spatial (image) domain and serves as input to the subsequent layers of the neural network. 
By design, subsequent layers of our proposed network takes real-valued inputs. 
As a result we stack the real and imaginary components of $\rvh$ as two separate channels.
The resulting tensor $\rvh_{R}$ is of size $\mathbb{R}^{r \times c \times 2}$, which is supplied as input to the downstream layers of the neural network. 
In practice, much like in real convolution neural networks, we convolve the \textit{k}-space input $\rvx$ with $p$ independent kernels $\{\rvz^1, \rvz^2, \ldots, \rvz^p \}$ to extract different features from the input, resulting in feature maps of size $\rvh_R \in \mathbb{R}^{p \times r \times c \times 2}$, which are supplied as input to the subsequent layers of the neural network.

Following the \textit{k}-space layer, we can adopt any standard architecture for the subsequent layers. The real-valued feature map $\rvh_R \in \mathbb{R}^{p \times r \times c \times 2}$ from the $k$-space layer is used as input to the subsequent layers, where instead of a $3$ channel input for \textsc{rgb} images, we have $p \times 2$ input channels. In this work, we use a Preact-ResNet \citep{he2016identity} for the subsequent layers. The output of this layer is a feature representation $\rvz \in \mathbb{R}^{h_{z}}$. This feature representation is used as input to a feed-forward network to output the probabilities of the positive and negative classes. Figure \ref{fig:model_arch} depicts the full architecture which we call the \textsc{kspace-net}. 
We can easily extend \textsc{kspace-net} to predict multiple pathologies from the same input. For each pathology, we can use a different feed-forward network with the feature representation $\rvz$ as input to each. 

Lastly, extending the \textsc{kspace-net} to work with under-sampled data is straightforward. We simply replace the full \textit{k}-space input $\rvx$ to the model with the under-sampled input $\rvx_\rvs$, which is obtained by taking an element-wise dot product with the sampling mask matrix $\rvs$: $\rvx_\rvs = \rvx \circ \rvs$ (see Section \ref{sec:background}). 

\section{End-to-End MR Triaging: \textsc{emrt} }\label{sec:emrt}
We now introduce {\bf E}nd-to-End {\bf MR} {\bf T}riaging (\textsc{emrt}): a novel method that infers the presence/absence of a disease (a binary decision) directly from a drastically small amount of \textit{k}-space data, skipping the image reconstruction process. 
The underlying motivating hypothesis behind \textsc{emrt} is that we can accurately infer the presence of a disease (a binary decision) from a small amount of carefully selected \textit{k}-space measurements so long as we are not concerned with reconstructing high-fidelity images. 
Towards that end, at a high-level, \textsc{emrt} learns to identify the subsets of the \textit{k}-space that have the largest predictive signal pertaining to the disease being identified without considering the quality of the image that would be generated using the data from identified subset. 
This is in contrast to the image reconstruction approaches where the requirement to generate a high quality image of the entire anatomy necessitates the sampling of a large portion of the \textit{k}-space.
Once the subset is identified, only the data from the identified \textit{k}-space subset is used by a \textsc{dl} to directly generate the final answer. 
To the best of our knowledge, \textsc{emrt} is the first method to propose classification of a disease directly from a carefully chosen (learned) subset of the \textit{k}-space. 
More formally \textsc{emrt} is a two-step algorithm. 

\begin{itemize}
    \item[Step 1:]
    In this step \textsc{emrt} searches for a subset of the \textit{k}-space that has the strongest signal that can help accurately infer the presence of the disease. 
    This is accomplished by learning a sparse sampling pattern $\rvs^*$, such that it maximizes the mutual information between under-sampled \textit{k}-space matrix $\rvx_{\rvs^*}$ and the response variable $\rvy$ (a binary variable indicating the presence/absence of the disease). 
    
    \item[Step 2:] 
    Once the sampling pattern $\rvs^*$ is learned, the second step involves using a \textsc{kspace-net} classifier (Section \ref{sec:classifier}) that takes as input the under-sampled \textit{k}-space data matrix $\rvx_{\rvs^*}$ to infer the presence of the disease $\rvy$, without reconstructing an intermediate image. 
\end{itemize}

To execute the above steps we need to answer the following questions, which we address in the following sub-sections: 
{\bf Q1.}  How to learn a sparse sampling pattern $\rvs^*$ of the \textit{k}-space matrix that maximizes the mutual information between the under-sampled \textit{k}-space $\rvx_{\rvs^*}$ and the response variable $\rvy$?; 
{\bf Q2.} How to train the \textsc{kspace-net} classifier that uses $\rvx_{\rvs^*}$ as input to accurately infer the disease $\rvy$.

\subsection{Learning the Sparse \textit{k}-Space Sampling Pattern}

\textsc{emrt} learns to identify a sampling pattern $\rvs^{*}$ over the \textit{k}-space matrix, such that the \textit{k}-space data $\rvx_{\rvs^*}$ corresponding to this pattern has the maximum information required to accurately infer the presence/absence of the disease. 
For any sampling pattern $\rvs$, \textsc{emrt} uses the mutual information between the output variable $\rvy$ and the corresponding under-sampled \textit{k}-space data $\rvx_{\rvs}$, as a surrogate for the information content in $\rvx_{\rvs}$ for disease inference.
Then for a given sampling rate $\alpha$, the process of identifying $\rvs^{*}$ (the optimal pattern) boils down to finding the sampling pattern that maximizes the mutual information between $\rvy$ and $\rvx_{\rvs^{*}}$.

Specifically, let $I(\rvy ; \rvx_{\rvs})$ denote the mutual information between $\rvy$ and $\rvx_{\rvs}$.
For a given sampling rate $\alpha$, \textsc{emrt} identifies a pattern $\rvs^{*}$, such that:
\begin{align}\label{eq:mi_optim}
  \rvs^* = \argmax_{\rvs \in {\{0, 1\}}^{r \times c}} I(\rvy \mid
  \rvx_{\rvs}),
\end{align}
where $\alpha = \frac{r \times c}{\norm{\rvs}_0}$ and $\rvs$ is the binary mask matrix of dimensions $r \times c$.
The mutual information $I(\rvy \mid \rvx_{\rvs})$ \citep{cover1999elements} is defined as:
\begin{align} 
  I(\rvy ; \rvx_{\rvs}) &= \E_{\rvx_{\rvs}} \kl{p(\rvy \mid \rvx_{\rvs}) \mid
  \mid p(\rvy)} \\
  &= \E_{\rvx_{\rvs}}  \E_{\rvy \mid \rvx_{\rvs}} \log p(\rvy \mid \rvx_{\rvs}) - \log
    p(\rvy) 
    \\ 
  &= \E_{\rvx_{\rvs}}  \E_{\rvy \mid \rvx_{\rvs}} \log p(\rvy \mid \rvx_{\rvs}) - C,
  \label{eq:mutual_information}
\end{align}
where $C$ is a constant independent of the sampling pattern 
$\rvs$, and $p(\rvy \mid \rvx_{\rvs})$ and $p(\rvy)$ are the conditional and the marginal distribution of the response variable respectively.
According to equation \ref{eq:mutual_information}, we can estimate the mutual information $I(\rvy \mid \rvx_{\rvs})$ if we are able to estimate the value of $p(\rvy \mid \rvx_{\rvs})$.
Since, we do not have access to the true conditional distribution $p(\rvy \mid \rvx_{\rvs})$, we can approximate the expected conditional log-likelihood by learning a probabilistic model $q(\rvy \mid \rvx_{\rvs}; \lambda)$ parameterized by $\lambda$.
However, learning a model for every sampling pattern $\rvs$ is infeasible even for moderately high dimensions.
To address this issue we draw upon the works of \cite{covert2020explaining,jethani2021have}, where the authors show that at optimality, a single model $q_{\text{val}}(\rvy \mid \rvx_{\rvs}; \lambda)$ trained with independently generated sampling patterns that are drawn independent of the data $\rvx, \rvy$, is equivalent to a conditional distribution of $\rvy$ for each sampling pattern. This approach is in contrast to approaches that explicitly model $\rvx$ \citep{sudarshan2020deep} and has been used in other applications \citep{jethani2022fastshap}.
As such, we train a model $q_{\text{val}}$ by minimizing the following loss function:

\begin{align*}
  \mathcal{L}(\lambda) = -\E_{\rvx, \rvy} \E_{\rvs \sim \pi}
  \log q_{\text{val}}(\rvy \mid
  \rvx_{\rvs}; \lambda),
\end{align*}
where $\pi$ is a distribution over the sampling pattern that is independent of the data $\rvx, \rvy$.
In \textsc{emrt}, distribution $\pi$ is a one-dimensional distribution and the \textsc{kspace-net} 
model (Section \ref{sec:classifier}) is used as $q_{\text{val}}$. The under-sampled data $\rvx_{\rvs}$
is created by masking the fully-sampled matrix $\rvx$ with a mask $\rvs \in \{0, 1\}^{r \times c}$. This masking ensures that the same model can be used as the input's dimensions are fixed.
This process is summarized in Algorithm  \ref{alg:qval_training}.

\begin{algorithm}[t]
  \caption{Estimating the conditional likelihood $q_{\text{val}}(\rvy \mid \rvx_{\rvs})$}
  \label{alg:qval_training}
\begin{algorithmic}
\STATE{{\bfseries Input:} Training data set $\mathcal{D}_{\text{tr}} = {\{(\rvx^i, \rvy^i)\}}_{i=1}^{N_{\text{tr}}}$,  model $q_{\text{val}}(\rvy \mid \rvx; \lambda)$ with initial parameters $\lambda$, mini-batch size $M$, acceleration factor $\alpha$, and prior distribution $\pi$ over sampling patterns}
\STATE{\textbf{Return:} Trained model $q_{\text{val}}(\rvy \mid \rvx_{\rvs};   \lambda^*)$}
\WHILE{not converged}
\STATE{Sample a mini-batch of training points of size $M$}
\STATE{Draw a sampling pattern $\rvs \sim \pi$, such that $\frac{r \times c}{\norm{\rvs}_0} = \alpha$}
\STATE{Update the model parameters
    \begin{align*}
                \lambda_{t+1} = \lambda_{t} + \frac{\gamma}{M} \sum_{i=1}^{M} \nabla_{\lambda}
                \log q_{\text{val}}(\rvy^i| \rvx^i_{\rvs}; \lambda_t)
    \end{align*}}
  \ENDWHILE{}
\STATE{Return the trained model $q_{\text{val}}(\rvy \mid \rvx_{\rvs}; \lambda^*)$}
\end{algorithmic}
\end{algorithm}

After training $q_{\text{val}}$, \textsc{emrt} uses it to define a scoring function $V: {\{0, 1\}}^{r \times c} \rightarrow \R$, for each sampling pattern $\rvs$ that estimates the mutual information between that subset of the k-space up to a constant \cref{eq:mutual_information}. 
Specifically,
\begin{equation}\label{eq:val_fn}
  V(\rvs) = \E_{\rvx} \E_{\rvy \mid \rvx} \log q_{\text{val}}(\rvy \mid
  \rvx_{\rvs}; \lambda) .
\end{equation}
The higher the score achieved by an sampling pattern the higher its diagnostic signal.
Therefore the objective of Equation \ref{eq:mi_optim} can be rewritten as
\begin{equation}
  \rvs^* = \argmax_{\rvs \in {\{0, 1\}}^{r \times c}} V(\rvs) , \text{ with } \frac{r \times c}{\norm{\rvs}_0} = \alpha .
\end{equation}

In practice $\rvs^*$ is approximated by a Monte Carlo search within the space of all sampling patterns.
$N$ candidate sampling patterns are drawn from the prior distribution $\pi$. 
Each drawn pattern is scored by the scoring function $V$ and the pattern with the highest score is selected as $\rvs^*$.
The details of the full algorithm are provided in Algorithm \ref{alg:learning_to_undersample}.

\subsection{Training the Direct \textit{k}-Space Classifier}
For inference during test time, we use the \textsc{kspace-net} classifier $q_{\text{val}}(\rvy \mid \rvx_{\rvs^*}; \lambda^*)$, trained using Algorithm \ref{alg:qval_training}, along with the optimized sampling pattern $\rvs^*$. 
As specified in Algorithm \ref{alg:qval_training}, during the training of this classifier, for every mini-batch we randomly sample a different sampling pattern from the distribution $\pi$. 
Through our experiments, we found that this is in-fact the key to training a reliable classifier.
We also explored retraining a classifier using data $\rvx_{\rvs^*}$, obtained from a fixed classification optimized sampling pattern $\rvs^*$. 
We compare these two approaches in \cref{sec:experiments}.
To summarize, the classifier $q_{\text{val}}(\rvy \mid \rvx_{\rvs}; \lambda^*)$ is trained with randomly sampled under-sampling patterns, however at test time we make inferences with a fixed under-sampling pattern.

\begin{algorithm}[t]
  \caption{Learning the sampling pattern $\rvs^*$}
  \label{alg:learning_to_undersample}
\begin{algorithmic}
\STATE{{\bfseries Input:} Validation data set $\mathcal{D}_{\text{val}} = {\{(\rvx^i, \rvy^i)\}}_{i=1}^{N_{\text{val}}}$, model $q_{\text{val}}(\rvy \mid \rvx; \lambda^*)$ with parameters $\lambda$, acceleration factor $\alpha$, number of candidate sampling patterns to generate $N$, and prior distribution $\pi$ over the sampling patterns}
\STATE{\textbf{Return:} Sampling pattern $\rvs^*$}
  \FOR{$j \in \{1, \dots, N\}$}
  \STATE{Sample $\rvs_j \sim \pi$ such that $\frac{r \times c}{\norm{\rvs}_0} = \alpha$}
  \STATE{Estimate the mutual information score in \cref{eq:val_fn} as follows
  \begin{align}
    \label{eq:val_fn_estimate}
    \widehat{V}(\rvs_j) = \frac{1}{N_{\text{val}}} \sum_{i=1}^{N_\text{val}} \log q_{\text{val}}(\rvy^{i} \mid \rvx_{\rvs_j}^{i}; \lambda^*)
  \end{align}
}
\ENDFOR{}
\STATE{Let $\rvs^* = \argmax_{j \in \{1, \dots, N\}} \widehat{V}(\rvs_j)$}
\end{algorithmic}
\end{algorithm}

\section{Experiments}\label{sec:experiments}
We evaluate the efficacy of \textsc{emrt} by comparing its performance to several benchmark models across multiple clinical tasks. 
Our experiments are structured to answer the following questions in order. 
{\bf Q1.} Can we infer the presence/absence of the disease directly from the \kspace data as accurately as the state-of-the-art image-based model trained on images reconstructed from the full k-space data?
{\bf Q2.} Using \textsc{emrt}, how much can we under-sample the \kspace input before we start to lose disease inference accuracy in comparison to the state-of-the-art image-based model trained on images reconstructed from the full \textsc{k}-space data? 
{\bf Q3.} For the same under-sampling factor, how much better (or worse) is the disease inference accuracy of the \textsc{emrt} model in comparison to the image-based model trained on images reconstructed from the under-sampled \kspace data using state-of-the-art image reconstruction method? 
{\bf Q4.} Is there any benefit of learning the sampling pattern using \textsc{emrt} that seeks to maximize the disease inference signal as compared to the sampling patterns proposed in the literature that optimize accurate image reconstruction or any heuristic based sampling pattern?

\subsection{Datasets}
Efficacy of \textsc{emrt} is assessed by comparing its performance to a variety of benchmark models on multiple clinical tasks across multiple anatomies. 
In particular we train and test our models to identify pathologies for three anatomies, namely knee \textsc{mr} scans,  brain \textsc{mr} scans, and abdominal \textsc{mr} scans. 
See Table \ref{table:data} for the description of data statistics for each of the three anatomies. 

\textbf{Knee \textsc{mr} Scans.}
We use \kspace data of the \textsc{mr} scans of the knees provided as part of the \textsc{fastmri} dataset \citep{zbontar2018fastmri} along with slice level annotations provided by the \textsc{fastmri}$+$ dataset \citep{zhao2021fastmri+}.
The dataset consists of multi-coil and single-coil coronal proton-density weighting scans, with and without fat suppression, acquired at the NYU Langone Health hospital system. Further sequence details are available in \cite{zbontar2018fastmri}.
The training, validation, and test sets consist of $816$, $176$, and $175$ volumes respectively.
The clinical task we solve is to predict whether a two-dimensional slice has a Meniscal Tear and/or an \textsc{acl} Sprain.

\textbf{Brain \textsc{mr} Scans.} 
We use the annotated slices of the \textsc{mr} scans of the brain also provided by the \textsc{fastmri} dataset \citep{zbontar2018fastmri} and then obtain the k-space data for these annotated slices using the \textsc{fastmri}+ dataset \citep{zhao2021fastmri+}.
A total of $1001$ volumes were annotated in the \textsc{fastmri}+ dataset out of a total of $5847$ volumes that were present in the \textsc{fastmri} dataset.
Each brain examination included a multi-coil single axial series (either T2-weighted FLAIR, T1-weighted without contrast, or T1-weighted with contrast).
The training, validation, and test sets consist of $700$, $150$, \& $151$ volumes respectively. 
We predict whether a two-dimensional slice has Enlarged Ventricles and/or Mass (includes Mass and Extra-axial Mass as in \citep{zhao2021fastmri+}).

\textbf{Abdominal \textsc{mr} Scans.}
The clinical task for the abdominal \textsc{mr} scans is the identification of a clinically significant prostate cancer (\textsc{cs-pca}), which is defined as a lesion within the prostate for which a radiologist assigns a Prostate Imaging Reporting And Data system (\textsc{pi-rads}) score \citep{weinreb2016pi} of $3$ or more.
We use the retrospectively collected bi-parametric abdominal \textsc{mr} scans performed clinically at NYU Langone Health hospital system.
It consists of scans from $313$ subjects who were referred due to suspected prostate cancer.
The scans were performed on a $3$ Tesla Siemens scanner with a $30$-element body coil array.
Examinations included an axial \textsc{t$2$}-weighted \textsc{tse} and an axial diffusion-weighted \textsc{epi} sequence using B values of $50$ and $1000$.
For our experiments we only used the data obtained using the \textsc{t$2$}-weighted sequence. 
For each scan volume, a board-certified abdominal radiologist examined each slice to identify the presence of lesion and assigned a \textsc{pi-rad} score to it. 
A slice is said to have \textsc{cs-pca}, if there exists at least one lesion in it with a \textsc{pi-rads} score of $3$ or more. 
We split the data into $218, 48$ and $47$ volumes for the training, validation and test sets, respectively. 

During the splits we make sure that scans from the same patient appear only in one of the three splits. 

Since the data for these scans is acquired using multiple coils, following \cite{zbontar2018fastmri}, we emulate it to be coming from a single coil using the emulated single-coil (\textsc{esc}) method \cite{tygert2020simulating}.
This results in a single \kspace matrix that is provided as an input to \textsc{emrt}. 
The primary motivation behind doing this was simplicity on our way to prove our hypothesis. 
In future work we will propose models that work directly with the multi-coil data.

\begin{table*}[t]
    \centering
    \small
    \begin{tabular}{lccccc}
\toprule
& \multicolumn{2}{c}{Knee \textsc{mr}} & \multicolumn{1}{c}{Abdominal \textsc{mr}} & 
\multicolumn{2}{c}{Brain \textsc{mr}} \\
\midrule
  & Mensc. Tear & \textsc{acl} Sprain & \textsc{cs-pca} &
  Enlg. Ventricles & Mass 
      \\ \midrule 
      Train slices & 29100 ($11\%$)  & 29100 ($3.6\%$) & 6649 ($5\%$)  & 11002 ($1.61\%$) & 11002 ($1.98\%$) \\
      Val slices  & 6298 ($11\%$)  & 6298 ($2.4\%$) & 1431 ($4.5\%$) & 2362 ($1.52\%$) & 2362 ($2.03\%$) \\
      Test slices & 6281 ($11\%$) & 6281 ($3\%$) & 1462 ($6\%$)
      & 2366 ($2.58\%$) & 2366 ($2.70\%$) \\
      \bottomrule
    \end{tabular}
    \caption{\textit{\textbf{Dataset statistics:} Number of slices in 
    the training, validation and test splits for each 
    task. Numbers in bracket are the percentages of slices in which the disease is visible (positive examples).}}
    \label{table:data}
\end{table*}

\subsection{Exp 1: Disease Inference Directly from \kspace}
Our first set of experiments tests the feasibility of inferring a disease directly from the \kspace data by comparing the performance of \textsc{kspace-net} to a \textsc{dl} model that uses high-fidelity images as input. 
Towards that end, we train the \textsc{kspace-net} model to solve the binary task of inferring the presence/absence of the disease using the full \kspace matrix $\tilde{\rvx}^{mc}$ that is emulated to be coming from a single coil using the \textsc{esc} algorithm \cite{tygert2020simulating} as input.
Performance of the \textsc{kspace-net} model is compared against the image-based deep learning models trained to infer presence of the disease from images reconstructed using the \textsc{rss} method from full \kspace data acquired using multiple coils. 
We train a pre-activation ResNet-50 \cite{he2016identity} model using these  $\tilde{\rvm}^{\textsc{rss}}$ images as its input.
We call this model \textsc{model$_{\textsc{rss}}$}. 
Disease inference accuracy of these models is the best one can hope to achieve from an 
image-based model, because the images are reconstructed from the full \kspace data and the models are trained using a rigorous hyper-parameter search to find the best performing model configuration.

\begin{table*}[th]
    \centering
    \small
    \begin{tabular}{lccccc}
\toprule
& \multicolumn{2}{c}{Knee \textsc{auroc}} & \multicolumn{1}{c}{\textsc{cs-pca}  \textsc{auroc}} & 
\multicolumn{2}{c}{Brain \textsc{auroc}} \\
\midrule
  & Mensc. Tear & \textsc{acl} Sprain & \textsc{cs-pca} &
  Enlg. Ventricles & Mass 
      \\ \midrule 
      \textsc{kspace-net} & $93.4 \pm 0.7$ & $90.8 \pm 1.5$ & $84.1 \pm 0.4$ & $92.3 \pm 2.0$ & $91.5 \pm 1.0$ \\
      \textsc{model$_{\textsc{rss}}$} & $92.1 \pm 1.0$ & $90.6 \pm 1.01$ & $83.1 \pm 1.6$ & $93.8 \pm 1.3 $ & $88.4 \pm 5$ \\
      \bottomrule
    \end{tabular}
    \caption{\textit{\textbf{Disease inference directly from \kspace:}
The \textsc{auroc} of the \textsc{kspace-net} model in comparison to a \textsc{dl} model trained on high-fidelity images to infer the presence/absence of specific diseases. 
The results clearly show that it is indeed feasible to infer the disease directly from the \kspace data as accurately as an image-based classifier.}}
    \label{table:experiment_1}
\end{table*}

\Cref{table:experiment_1} provides the \textsc{auroc} of the \textsc{kspace-net} model in comparison to  \textsc{model$_{\textsc{rss}}$}.
The results clearly show that it is indeed feasible to infer the presence of the disease directly from the \kspace data as accurately as a finely tuned \textsc{dl} model trained on high-fidelity images. 
This result is not surprising, since transformation from \textit{k}-space to image space is achieved using \textsc{ifft}, which is a deterministic and lossless operation. 
What is surprising is that in some cases the \textsc{kspace-net} model performs better than the image-based model. 
While this question is left for future work, we conjecture that the reason behind this performance gap is that the \textsc{kspace-net} model uses as input the entire complex data where as the image-based model uses only the magnitude of the complex matrix in the image space (as is the widespread norm in medical image analysis). 
Lastly, these results are particularly impressive when one takes into account that the \textsc{kspace-net} model takes as input the data emulated from a single coil (which has a lower \textsc{snr}) whereas \textsc{model$_{\textsc{rss}}$} is using the full multi-coil data. 
As part of the future work we are working on extending the \textsc{kspace-net} model to ingest multi-coil data directly. 

\begin{figure}[t!]
\small
\begin{center}
\includegraphics[width=0.35\textwidth]{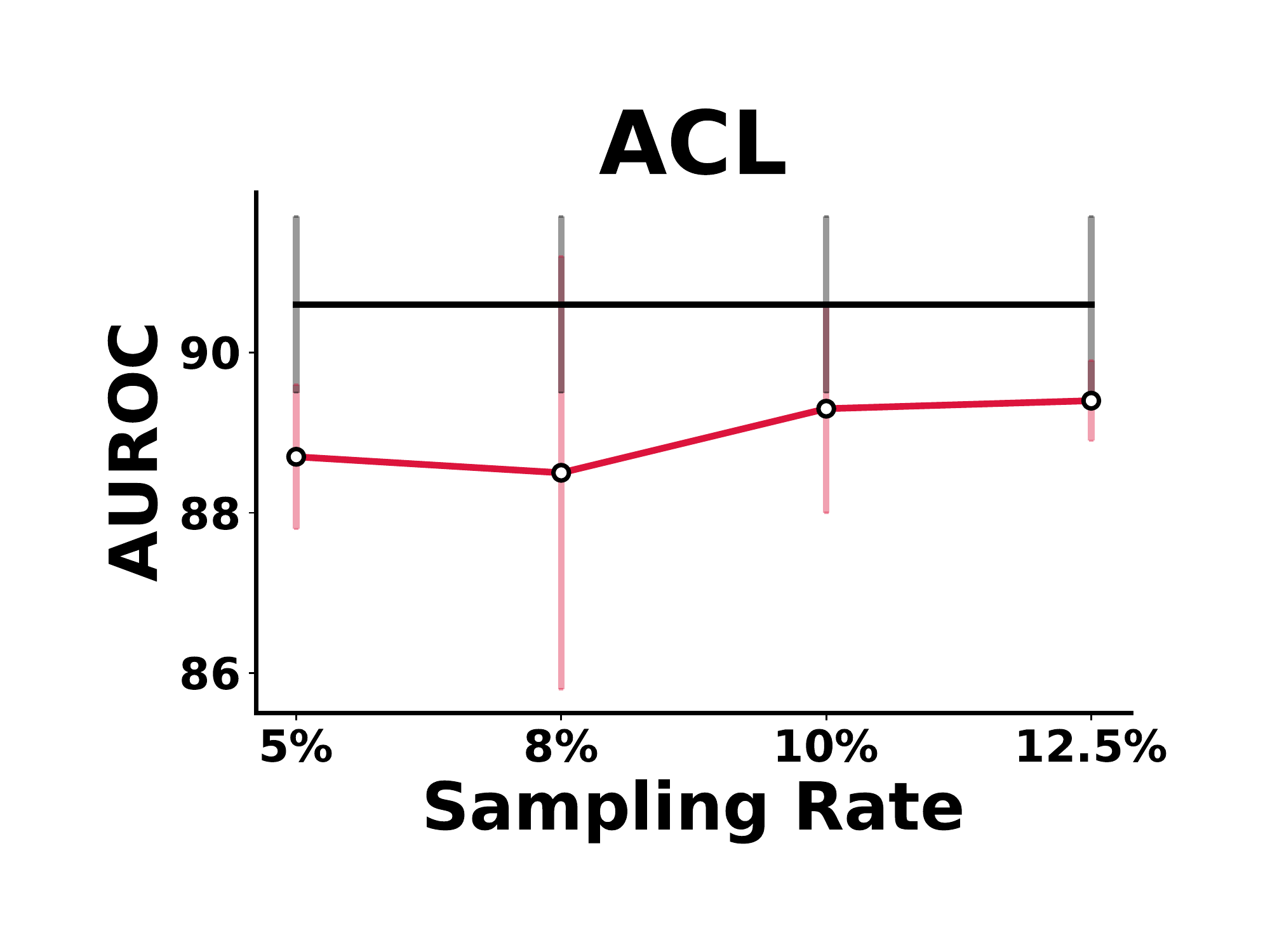}\includegraphics[width=0.35\textwidth]{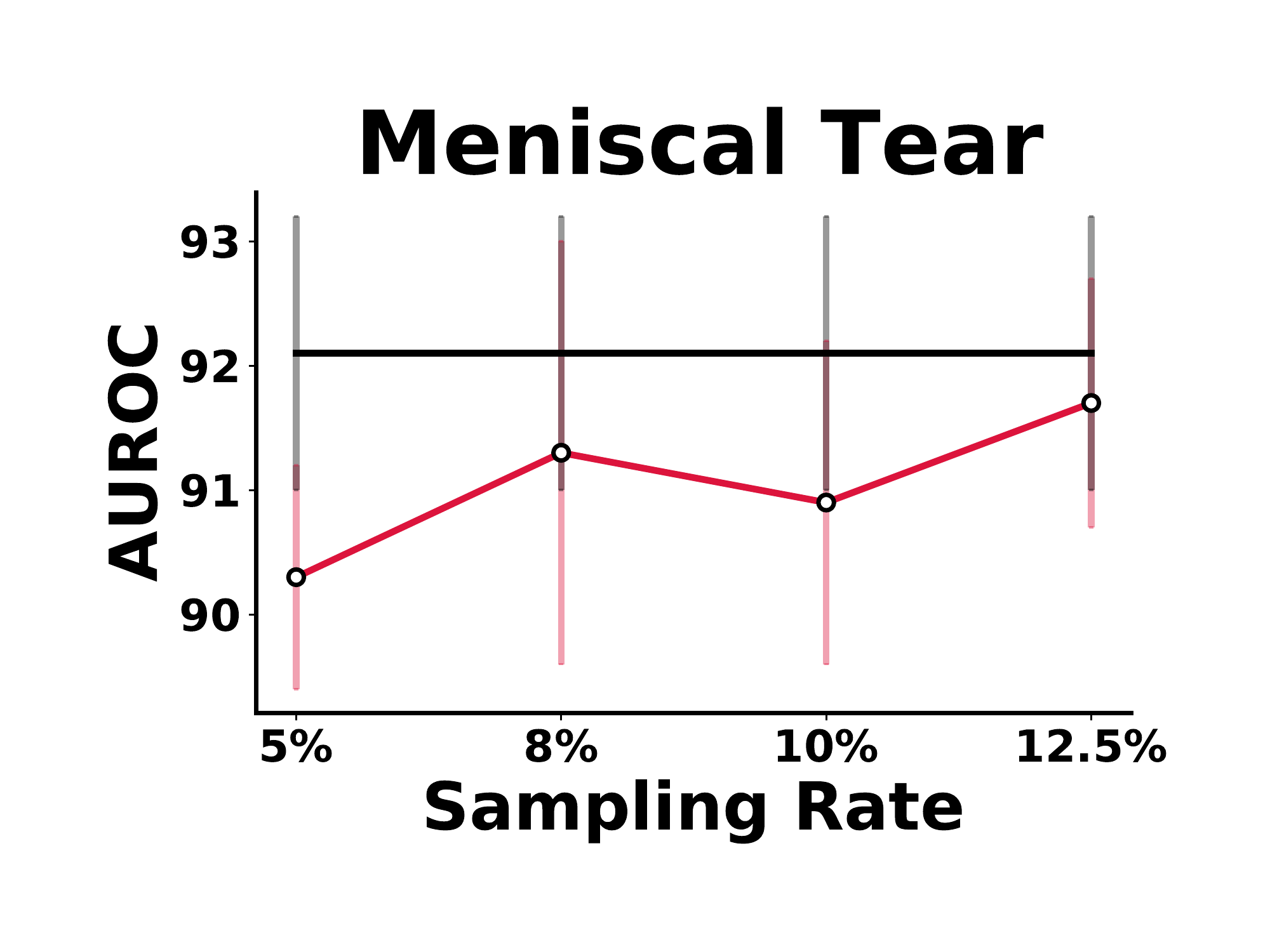}\includegraphics[width=0.35\textwidth]{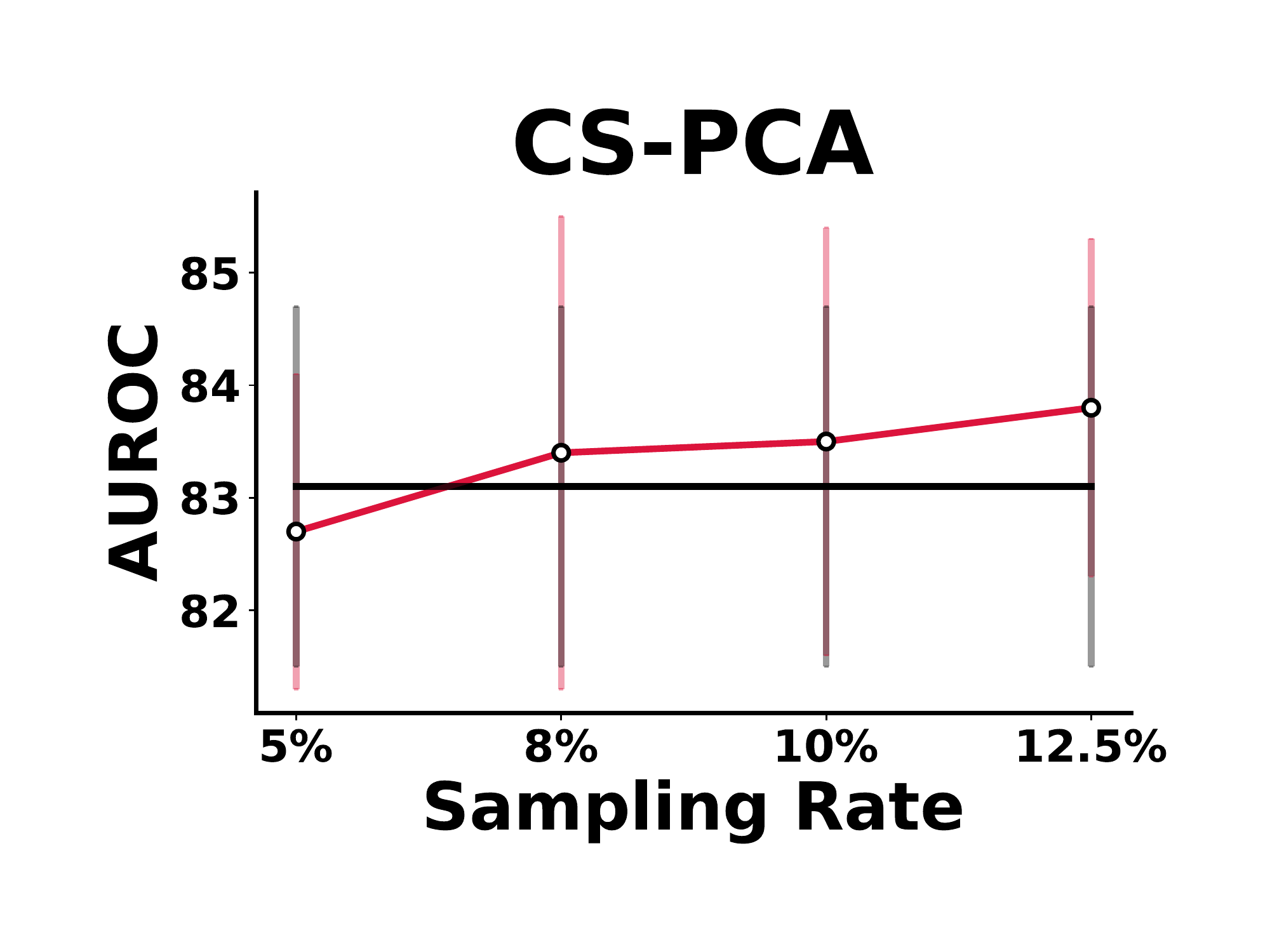}
\end{center}
\includegraphics[width=0.35\textwidth]{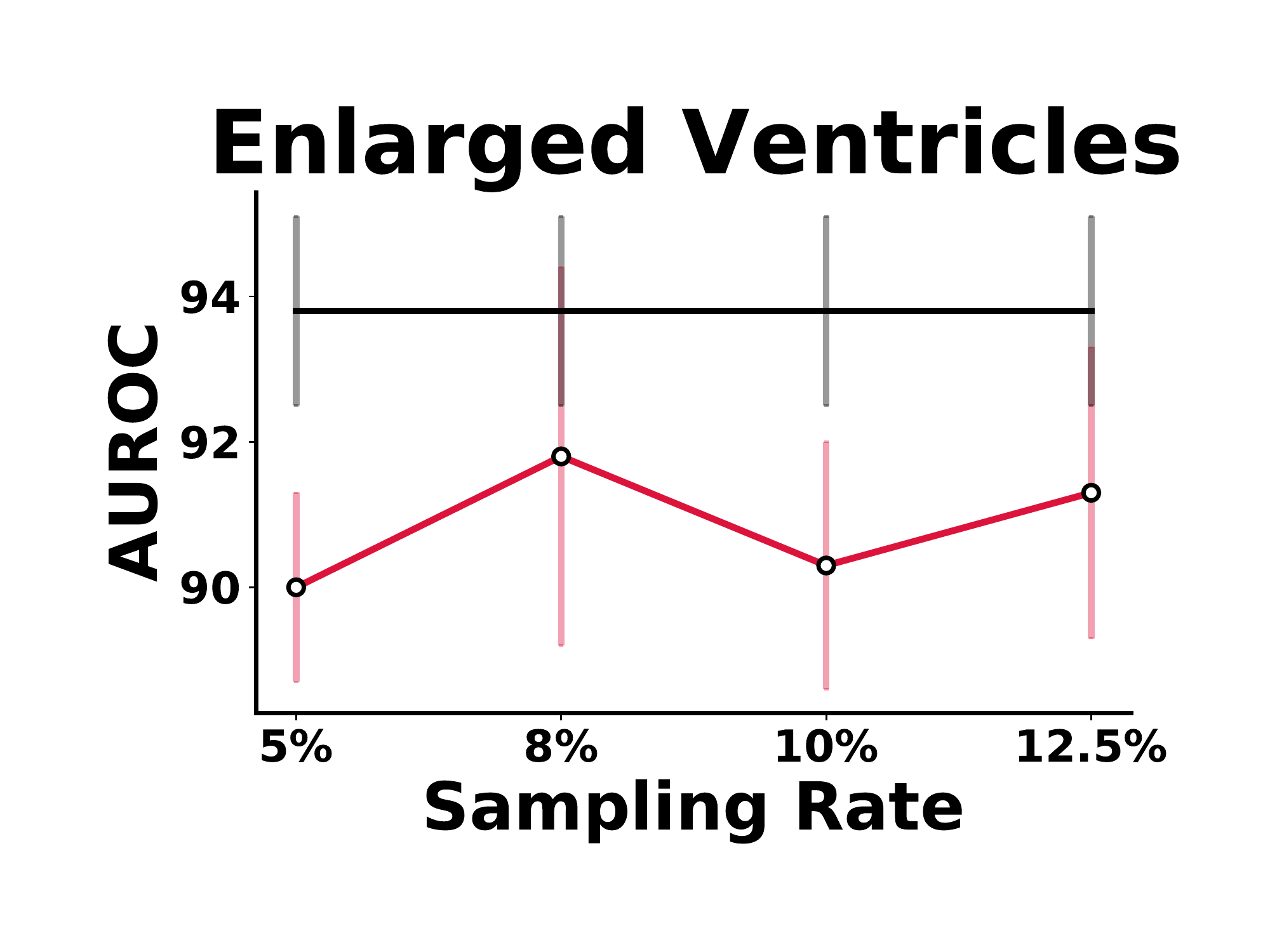}\includegraphics[width=0.35\textwidth]{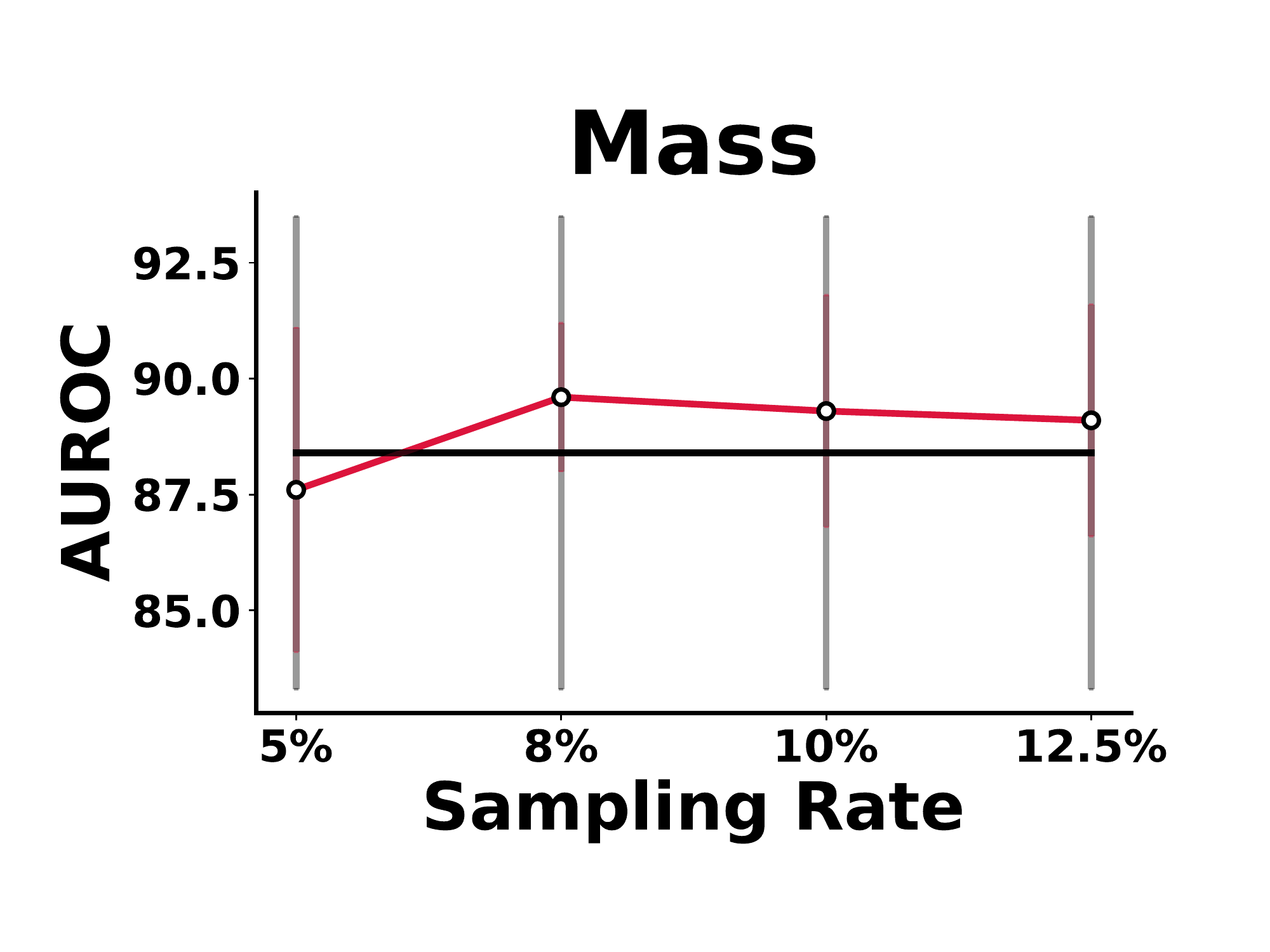}
\caption{\textit{\textbf{Performance of \textsc{emrt} against \textsc{model}$_{\textsc{rss}}$:} Top panel shows \textsc{auroc} on the test set of the \textcolor{red}{\textbf{\textsc{emrt} (red)}} at different sampling factors in comparison to the \textsc{auroc} of \textbf{\textsc{model}$_{\textsc{rss}}$ (black)} trained using the fully-sampled \textit{k}-space data. }}
\label{fig:class_results}
\end{figure}

\subsection{Exp 2: Exploring the Limits on Under-Sampling  the \kspace Using \textsc{emrt}}
In our second set of experiments, we estimate the extent to which one can under-sample the \kspace data and still infer the presence of the disease (using the \textsc{kspace-net} model) as accurately as an image-based classifier using high-fidelity images as input. 
We sample the \kspace at different sampling rates $\alpha$ ($\in \{$5$\%, $8$\%, $10$\%, $12.5$\%\}$) and train a \textsc{kspace-net} for each $\alpha$. 
For the given sampling rate $\alpha$, the sampling pattern is learnt using the \textsc{emrt} procedure, summarized in  \Cref{alg:qval_training} and \Cref{alg:learning_to_undersample}.

\Cref{fig:class_results} and \cref{table:experiment_emrt_v_rss} give the \textsc{auc}, Sensitivity, and Specificity of the \textsc{emrt} model at different sampling rates and compares its performance to the \textsc{model}$_{\textsc{rss}}$. 
We observe that at high sampling rates, the performance of \textsc{emrt}, in terms of \textsc{auc} and sensitivity-specificity, does not deteriorate significantly in comparison to the \textsc{dl} trained trained on high-fidelity images reconstructed using the full \kspace data. 
This experiment demonstrates that if the goal is to  simply infer the presence/absence of the disease, without the concern to reconstruct a high-fidelity image, then we can afford to significantly under-sample the \kspace data (as low as $5$\%) without any significant loss in performance. 
This is in contrast to \citep{muckley2021results}, which reports that in the Fast\textsc{mri} challenge, all submissions had reconstructed images that started missing clinically relevant pathologies at sampling rates less than $25\%$ of the data.
Figure \ref{fig:prostate_deterioration} shows the sequence of images reconstructed from the \kspace data corresponding to the sampling patterns learnt by \textsc{emrt}. One can clearly see that the pathology visible is the image reconstructed from the full \kspace is hard to discern in images generated from under-sampled data.
Furthermore, it becomes successively hard to identify the pathology as we decrease the amount of data used. 

\begin{table*}[th]
    \centering
    \small
    \begin{tabular}{lccccc}
\toprule
& \multicolumn{2}{c}{Knee \textsc{sens/spec}} & \multicolumn{1}{c}{\textsc{cs-pca} \textsc{sens/spec}} & 
\multicolumn{2}{c}{Brain \textsc{sens/spec}} \\
\midrule
  & Mensc. Tear & \textsc{acl} Sprain & \textsc{cs-pca} &
  Enlg. Ventricles & Mass 
      \\ \midrule 
      \textsc{emrt} & $81/83$ & $80/81$ & $88/65$ & $86/82$ & $89/70$ \\
      \textsc{model$_{\textsc{rss}}$} & $83/86$ & $81/82$ & $88/60$ & $78/94$ & $82/80$ \\
      \bottomrule
    \end{tabular}
    \caption{\textit{\textbf{Performance of \textsc{emrt} 
    against \textsc{model}$_{\textsc{rss}}$:} Test Sensitivity/Specificity of \textsc{emrt} and \textsc{model}$_{\textsc{rss}}$ obtained 
    using an operating point with $85\%$ Sensitivity on the validation set. 
The Sensitivity/Specificity results are reported using  a
 sampling factor $\alpha =5\%$ for knee \textsc{mr} and $8\%$ for
  brain and prostate \textsc{mr} scans. The input to \textsc{model}$_{\textsc{rss}}$ is the fully-sampled \kspace. See \cref{appsec:metrics} for confidence intervals.}}
    \label{table:experiment_emrt_v_rss}
\end{table*}

\begin{figure}[t!]
\small
\begin{center}
\includegraphics[width=0.35\textwidth]{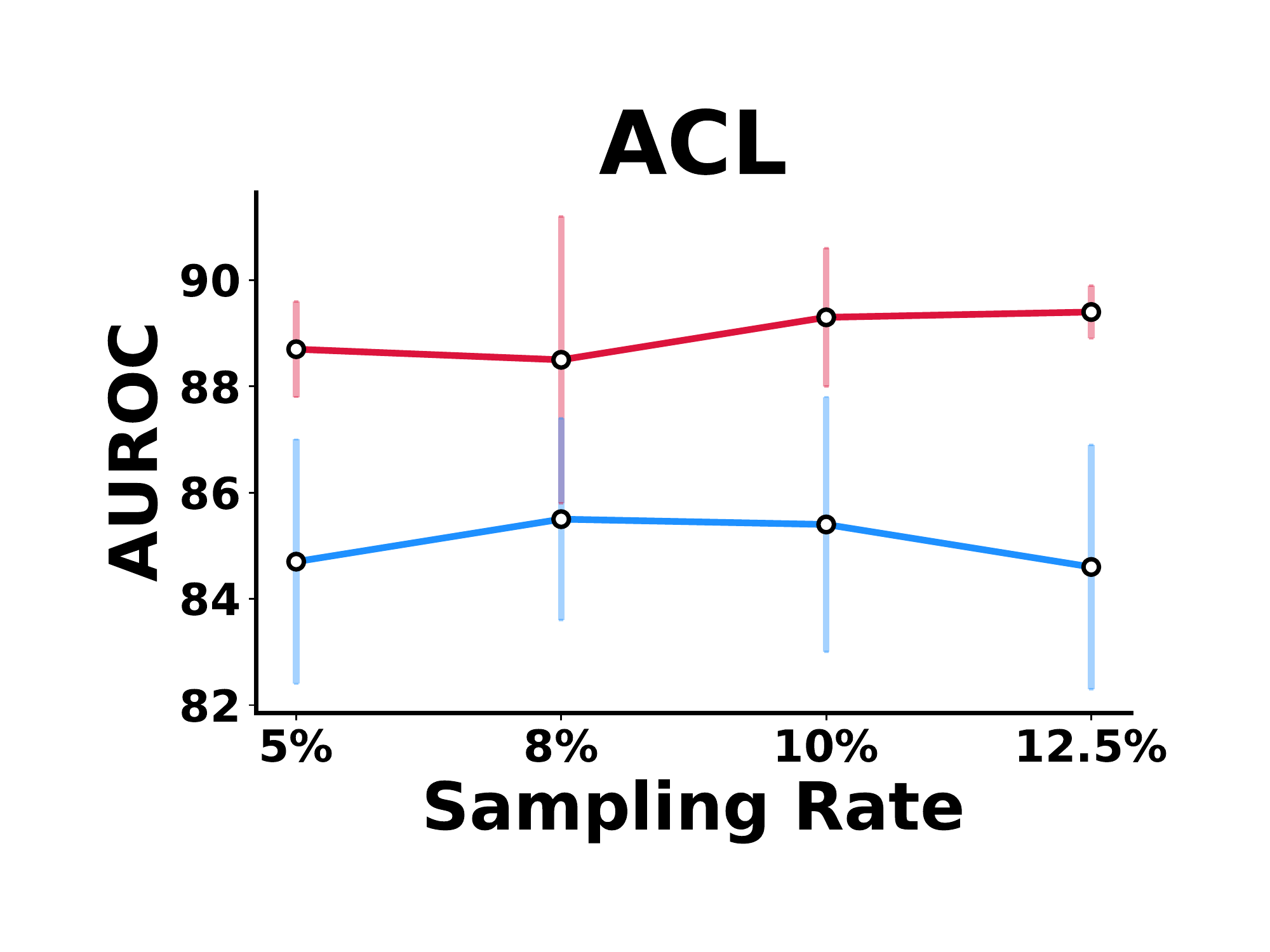}\includegraphics[width=0.35\textwidth]{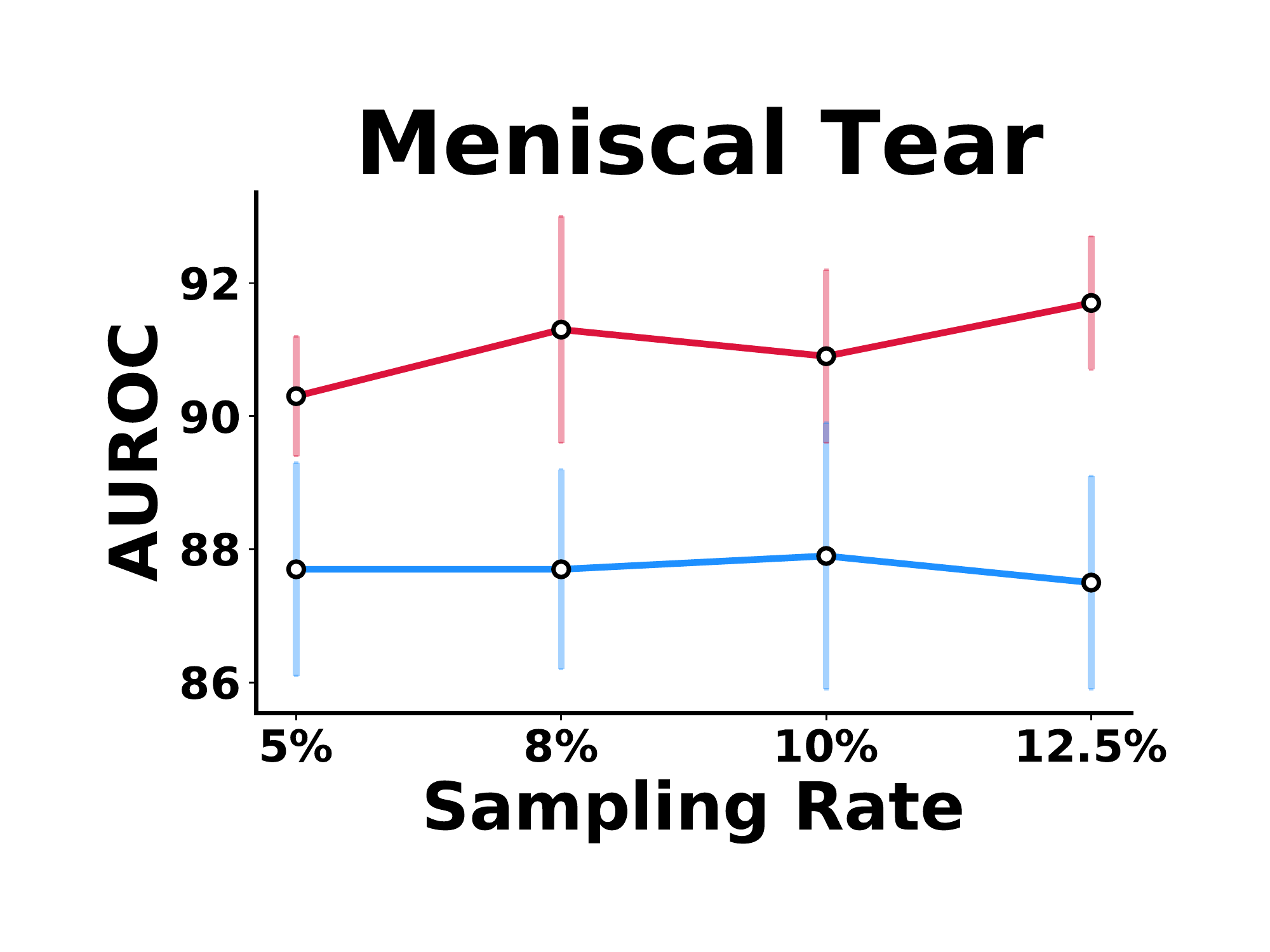}\includegraphics[width=0.35\textwidth]{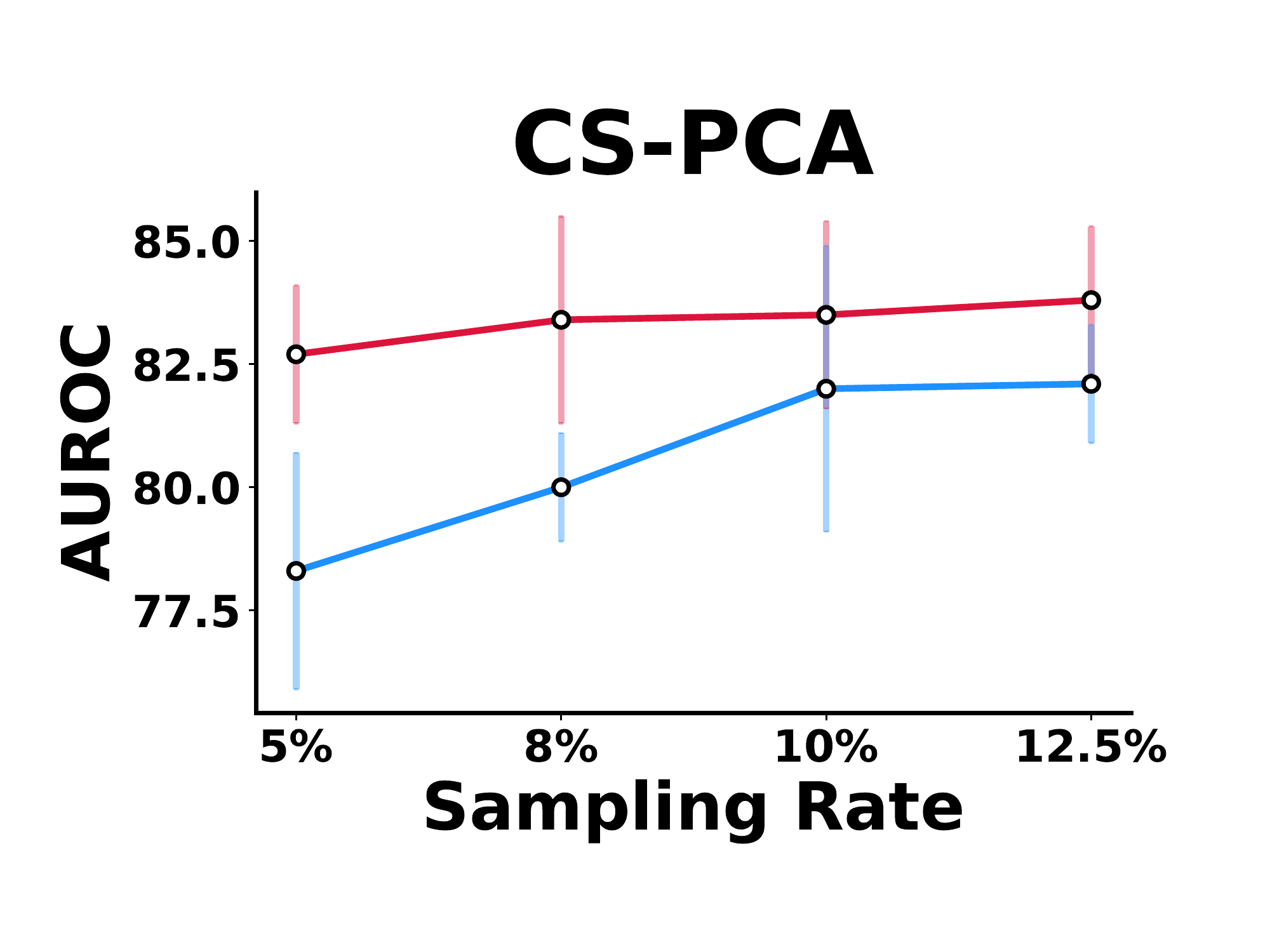}
\end{center}
\includegraphics[width=0.35\textwidth]{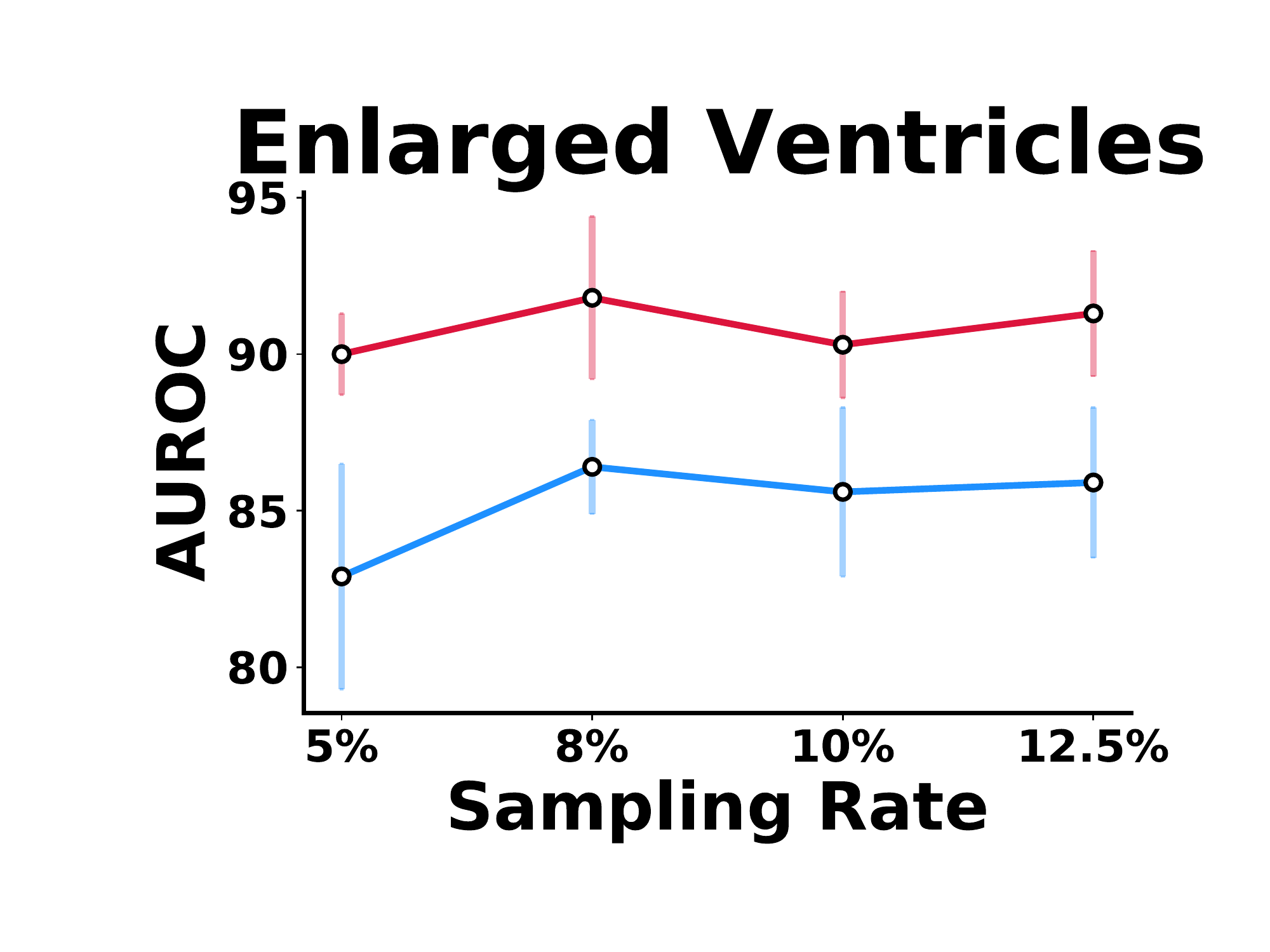}\includegraphics[width=0.35\textwidth]{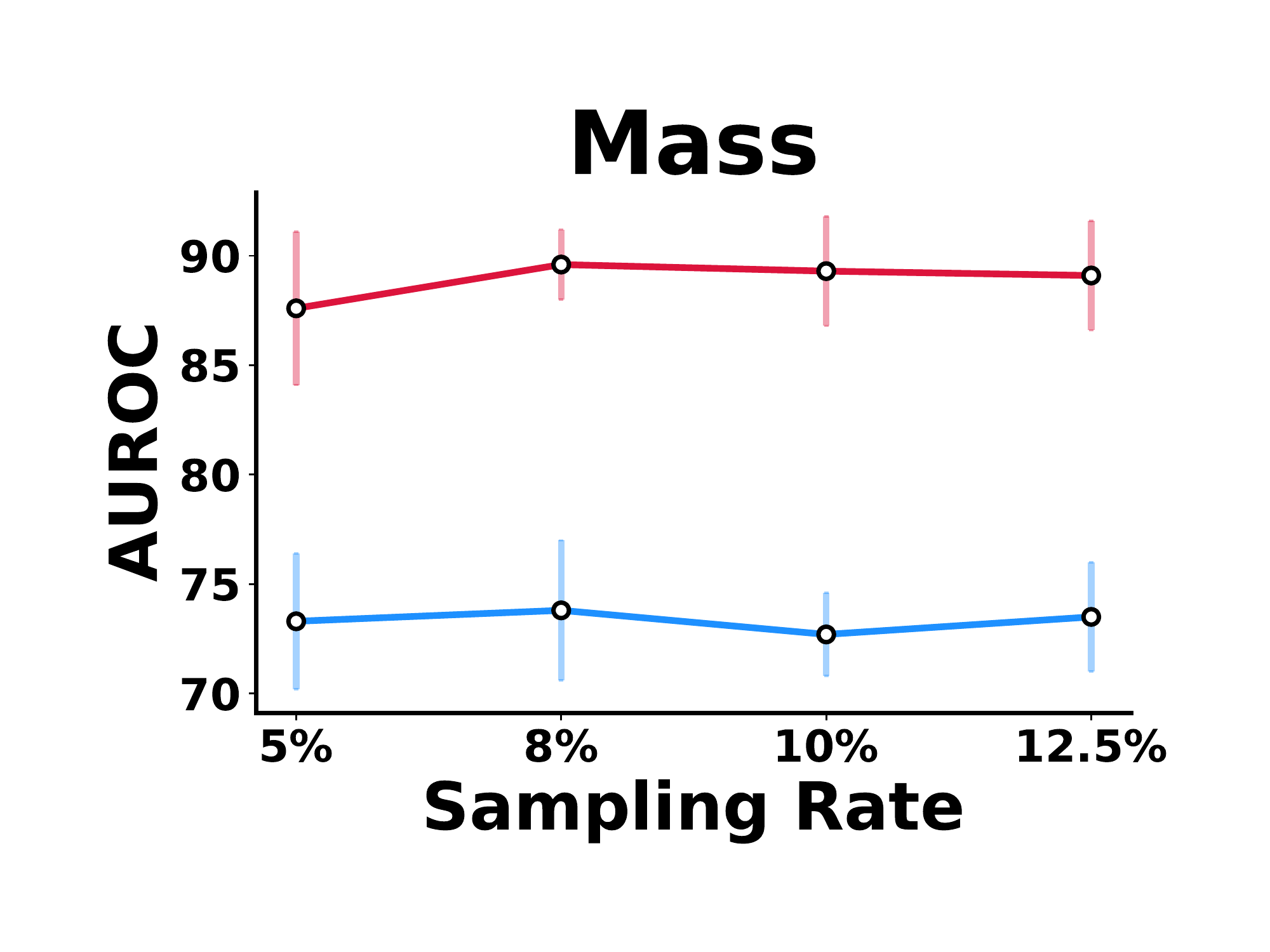}
\caption{\textit{\textbf{Performance of \textsc{emrt} against \textsc{model}$_{\textsc{loupe:varnet}}$:} Top panel shows \textsc{auroc} on the test set of the \textcolor{red}{\textbf{\textsc{emrt} (red)}} at different sampling factors in comparison to the \textsc{auroc} of \textcolor{blue}{\textsc{model}$_{\textsc{loupe:varnet}}$ (blue)}.  Note that for all pathologies, \textsc{mr} scans and all sampling rates $\alpha \in \{5\%, 8\%, 10\%, 12.5\% \}$ \textsc{emrt} outperforms \textsc{model}$_{\textsc{loupe:varnet}}$.}}
\label{fig:emrt_vs_recon}
\end{figure}

\begin{table*}[th]
    \centering
    \small
    \begin{tabular}{lccccc}
\toprule
& \multicolumn{2}{c}{Knee \textsc{sens/spec}} & \multicolumn{1}{c}{\textsc{cs-pca} \textsc{sens/spec}} & 
\multicolumn{2}{c}{Brain \textsc{sens/spec}} \\
\midrule
  & Mensc. Tear & \textsc{acl} Sprain & \textsc{cs-pca} &
  Enlg. Ventricles & Mass 
      \\ \midrule 
      \textsc{emrt} & 81/83 & $80/81$ & $88/65$ & $86/82$ & $89/70$ \\
      \textsc{model$_{\textsc{loupe:varnet}}$} & $81/79$ & $74/81$ & $86/54$ & $84/72$ & $74/56$ \\
      \bottomrule
    \end{tabular}
    \caption{\textit{\textbf{Performance of \textsc{emrt} against \textsc{model}$_{\textsc{loupe:varnet}}$:} Test Sensitivity/Specificity of \textsc{emrt} and \textsc{model}$_{\textsc{loupe:varnet}}$ obtained using an operating point with $85\%$ Sensitivity on the validation set. 
The Sensitivity/Specificity results are reported using  a sampling factor $\alpha =5\%$ for knee \textsc{mr} and $8\%$ for brain and prostate \textsc{mr} scans. 
See \cref{appsec:metrics} for confidence intervals.}}
    \label{table:experiment_emrt_v_varnet}
\end{table*}

\subsection{Exp 3: Reconstructed Images vs Direct \kspace When Under-Sampling}
So far we have established that we can infer the presence/absence of a disease directly from \kspace data.
In addition, when we are not concerned with reconstructing intermediate images, we only need a fraction of the \kspace data to infer the disease without compromising accuracy in comparison to a model trained on images reconstructed from the full \kspace data. 
When using under-sampled \kspace data however, another way to infer the disease presence is by first reconstructing an intermediate image from the under-sampled data and then training a classifier on these images to infer the disease. 
Our third set of experiments are structured to answer the following question: ``how is the disease inference accuracy impacted if we use a \textsc{dl} model trained on images reconstructed from the under-sampled \kspace data in comparison to the \textsc{emrt}, which infers the disease directly from the \kspace data?''

Towards that end, we compare the performance of \textsc{emrt} against the image-based classifiers which are trained using images reconstructed from the under-sampled \kspace data. 
For the image-based classifiers, the sampling pattern used is the one obtained by the \textsc{loupe} method \cite{bahadir2019learning}: a state-of-the-art method proposed in the literature which learns a sampling pattern over the \kspace such that the data corresponding to it gives the best possible reconstructed image. 
Furthermore, we use the state-of-the-art image reconstruction model, namely the \textsc{VarNet} model \citep{sriram2020end}, to reconstruct the images from the under-sampled \kspace data. 
We denote this benchmark by \textsc{model}$_{\textsc{loupe:varnet}}$, identifying the methods used for learning the sampling pattern and the method used to reconstruct the images from the learnt sampling pattern respectively. 

\Cref{fig:emrt_vs_recon} and \cref{table:experiment_emrt_v_varnet} compare the performance of the two sets of models. We observe that for all the abnormalities and for all sampling rates,  \textsc{emrt} outperforms \textsc{model}$_{\textsc{loupe:varnet}}$. 
The bottom panel of \Cref{fig:emrt_vs_recon} shows the sensitivity and specificity of the models obtained at $5$\% sampling rate for knees, and $8$\% sampling rate for abdomen and brain. 
For a given sensitivity, \textsc{emrt} has a significantly better specificity compared 
to \textsc{model}$_{\textsc{loupe:varnet}}$, translating to lower number of false positive cases.
Furthermore we observe that for some pathologies, such as \textsc{cs-pca} and Enlarged Ventricles, there is a
sharp decrease in the \textsc{auroc} compared to \textsc{emrt}, which for the most 
part remains stable across all sampling factors and for all the pathologies.
.

\begin{figure}[th]\label{fig:recon_fig}
\centering
\includegraphics[scale=0.3]{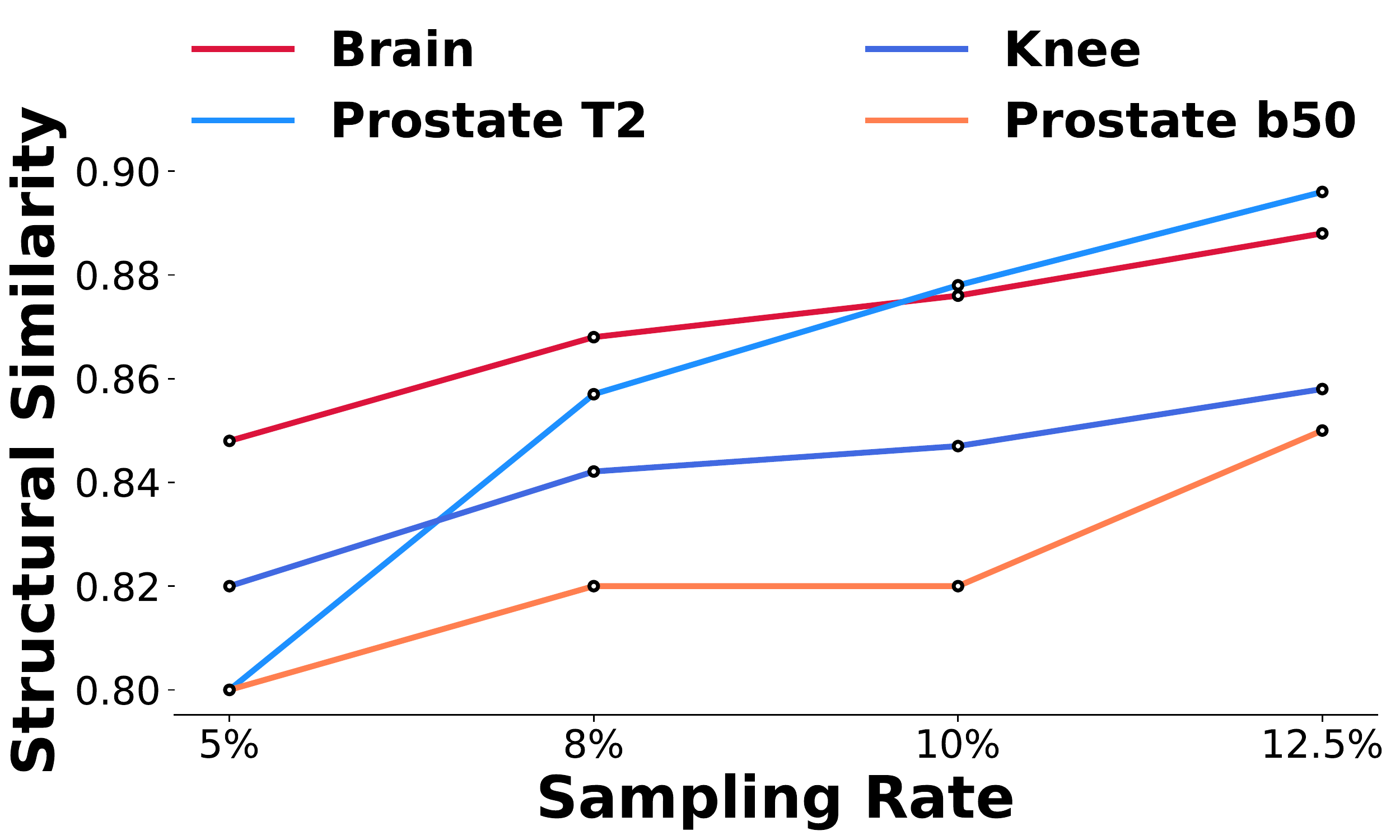}
\includegraphics[width=0.8\textwidth]{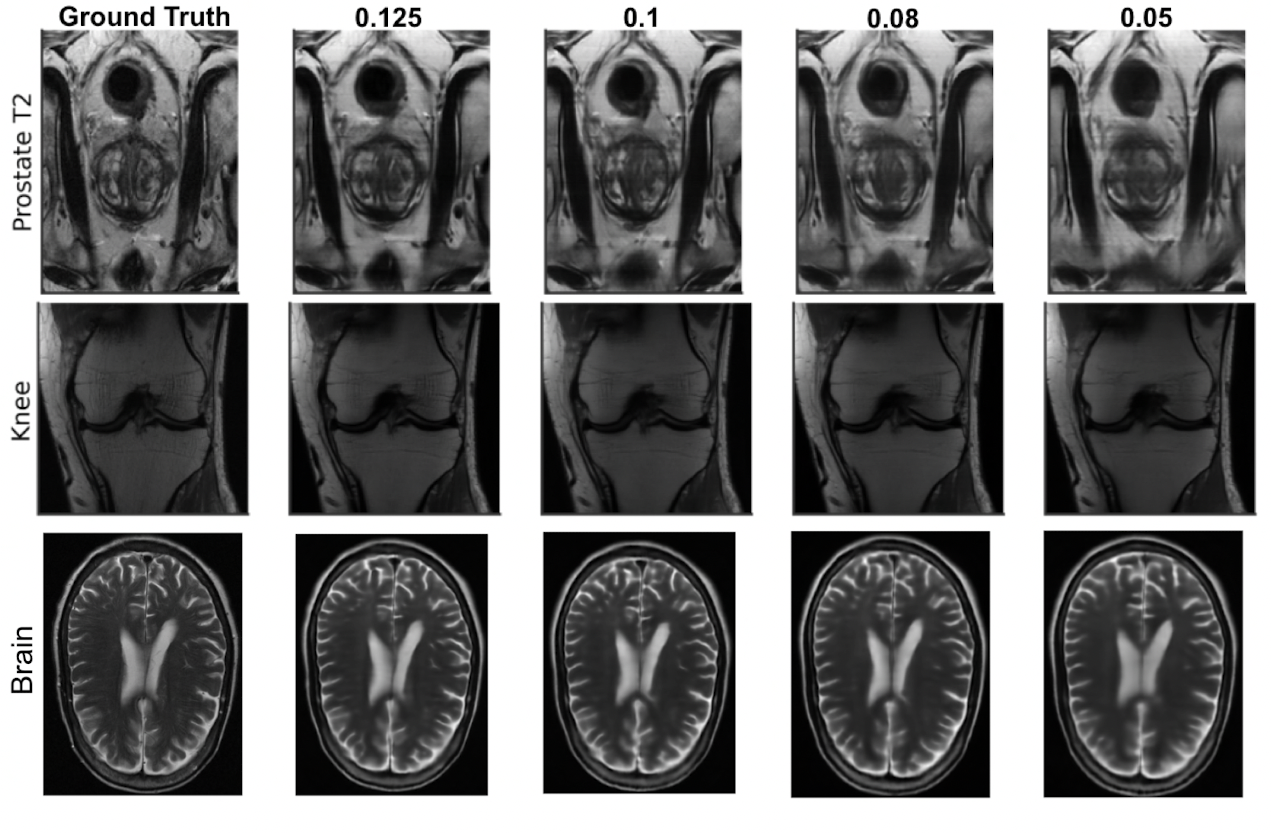}

\caption{\textit{\textbf{Performance of image reconstruction:}
Reconstruction methods
are an essential component of the indirect classification
benchmark.
In this figure, we plot the reconstruction performance of the
best performing reconstruction methods
  at increasing sampling rates
  $\alpha \in \{5\%, 8\%, 10\%, 12.5\%\}$. }}\label{fig:recon_metrics}
\end{figure}

Lastly, to validate the correctness of our implementation of the image reconstruction method (\textsc{VarNet} \citep{sriram2020end}) we also report the structural similarity (\textsc{ssim}) metric in \cref{fig:recon_metrics}, a commonly used metric to measure reconstruction quality. 
Our \textsc{ssim} numbers are within the ballpark of the state-of-the-art reported in literature. 
Specifically, for $12.5\%$ sampling rate, the knee reconstruction \textsc{ssim} is $0.82$ compared to $0.88$ reported in \citep{sriram2020end} and the brain reconstruction \textsc{ssim} is $0.89$ compared to $0.94$ reported in \citep{sriram2020end}.

\subsection{Exp 4: Benefits of Learning Sampling Pattern Using \textsc{emrt}}
In our next set of experiments we show two things.
First, we show that the sampling pattern learnt by \textsc{emrt} (which optimizes the classification accuracy) is different from the ones learnt by any method  that optimizes a reconstruction metric (such as \textsc{loupe}). 
Second, we show the benefits of learning a sampling pattern that explicitly optimizes the disease classification accuracy (as achieved by \textsc{emrt}) in comparison to other sampling pattern.

\begin{figure}[t]\centering
  \includegraphics[width=\textwidth]{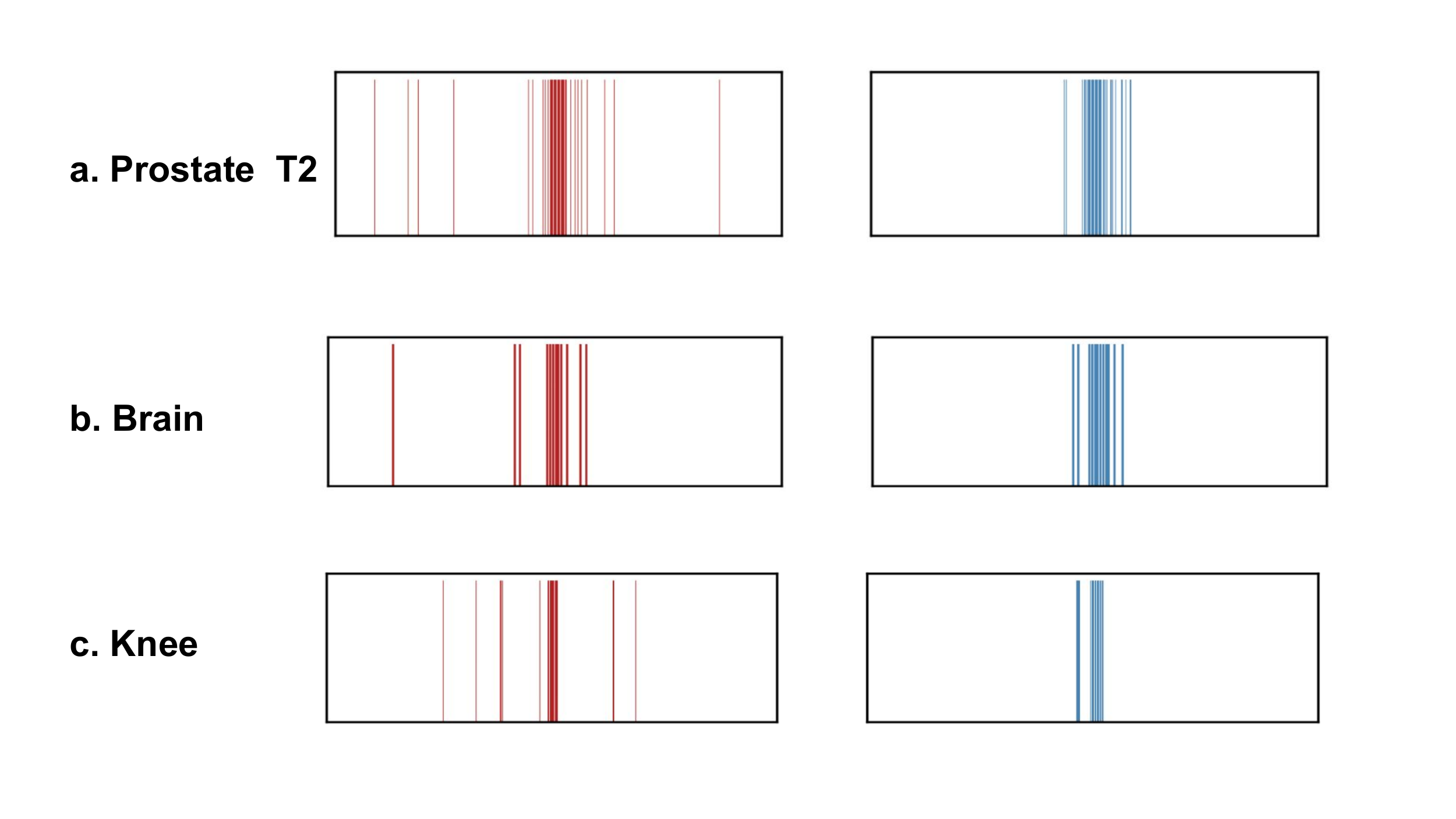}
  \caption{\textit{{\bf Contrasting sampling patterns:}
  Here we compare the  sampling patterns learnt by \textsc{emrt} that optimizes classification accuracy versus the patterns learnt by \textsc{loupe} that optimizes the reconstruction metric for different diseases.  
  \textsc{emrt} is learning a mix of low and high frequencies (red lines spread across the spectrum). 
  Whereas \textsc{loupe} predominantly is picking low frequencies (blue lines clustered around the center). The prostate and brain sampling patterns are sampled with $8\%$ sampling rate while knee \textsc{mr} patterns are sampled at a $5\%$ sampling rate.}}
  \label{fig:learnt_masks}
\end{figure}

\begin{figure}[th!]
  \small
  \begin{center}
  \includegraphics[width=0.5\textwidth]{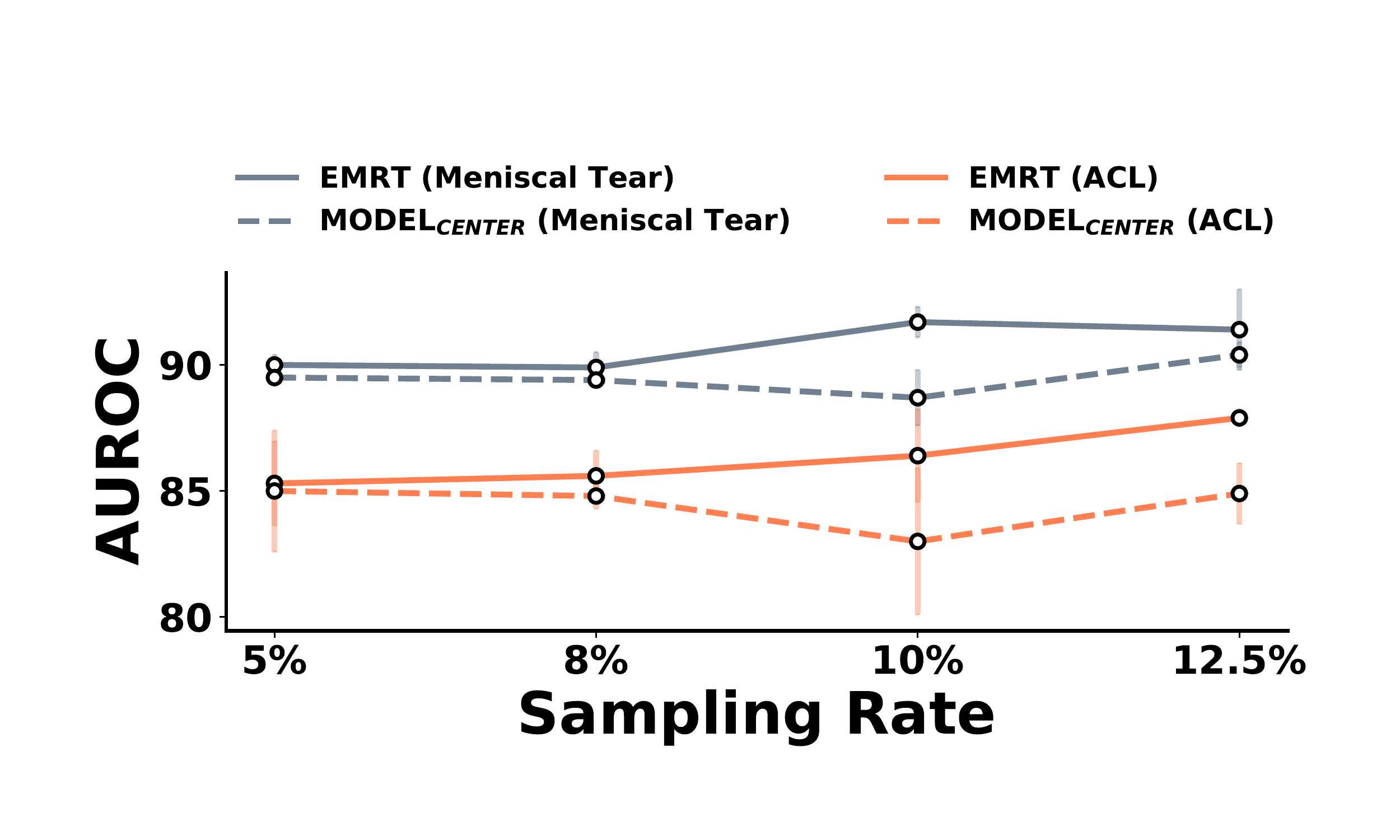}\includegraphics[width=0.5\textwidth]{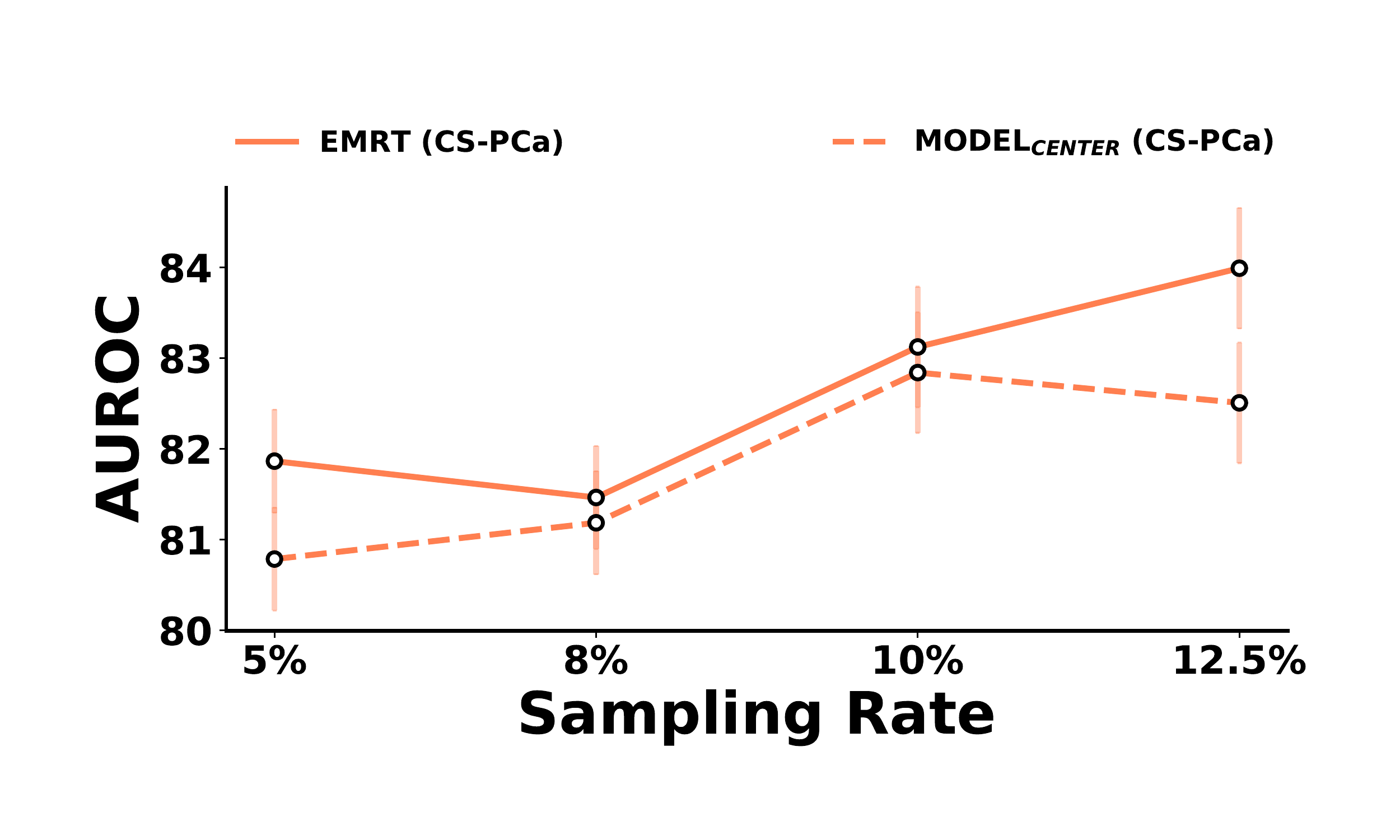}
  \end{center}
  \includegraphics[width=0.5\textwidth]{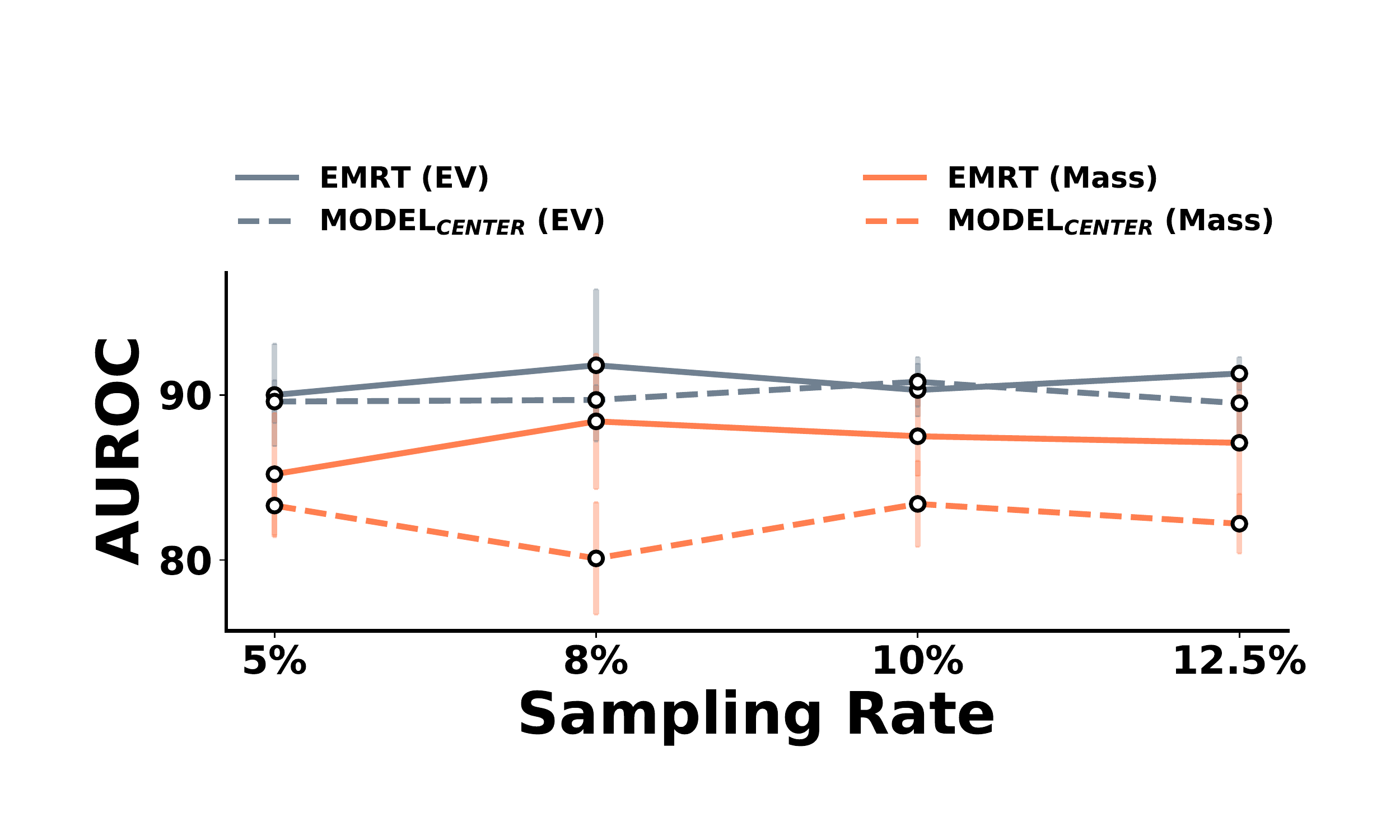}
  \caption{\textit{\textbf{Benefits of learning the sampling pattern:} 
  Figure shows \textsc{auroc} of \textsc{emrt} (which learns a sampling pattern that optimizes the disease classification accuracy) in comparison to the \textsc{auroc} of  \textsc{model}$_{\textsc{center}}$ which uses a fixed sampling pattern that is center-focused. 
  Superior performance of \textsc{emrt} across all tasks across all the sampling rates is indicative of the benefits of learning a sampling pattern that explicitly optimizes the classification accuracy.}}
  \label{fig:fixed_vs_center}
  \vspace{-0.1in}
  \end{figure}

\Cref{fig:learnt_masks} contrasts the classification optimized sampling pattern learnt by \textsc{emrt} versus the reconstruction-optimized sampling patterns learnt by \textsc{loupe}. 
We clearly see that the sampling pattern learnt by \textsc{emrt} is composed of a mixture of a set of low frequencies (red lines clustered around the center)  and a set of high frequencies (red lines spread away from the center). 
This is in contrast to the predominantly low frequencies selected by \textsc{loupe}, that are largely concentrated around the center.

Next, to show the benefits of learning a sampling pattern catered towards explicitly optimizing the disease identification accuracy, we compare the performance of \textsc{emrt} against another \textsc{kspace-net} model that is trained to identify the disease using a fixed sampling pattern consisting of only low frequencies (center-focused \textit{k}-space lines). 
We denote this model by \textsc{model}$_{\textsc{center}}$.
Figure \ref{fig:fixed_vs_center} compares the performance of the two sets of classifier.
As evident from the figure, performance of \textsc{emrt} is better than the performance of \textsc{model}$_{\textsc{center}}$ across all tasks, pointing towards the benefits of learning the sampling pattern that optimizes the classification accuracy. 
The performance gap is larger for tasks for which the frequencies learnt by \textsc{emrt} are more spread away from the center of the frequency spectrum, such as Mass in the brain scans and \textsc{cs-pca} in prostate scans.

\subsection{Exp 5: The Role of Random Subset Training in \textsc{emrt}}
\begin{figure}[t!]
\small
\begin{center}
\includegraphics[width=0.50\textwidth]{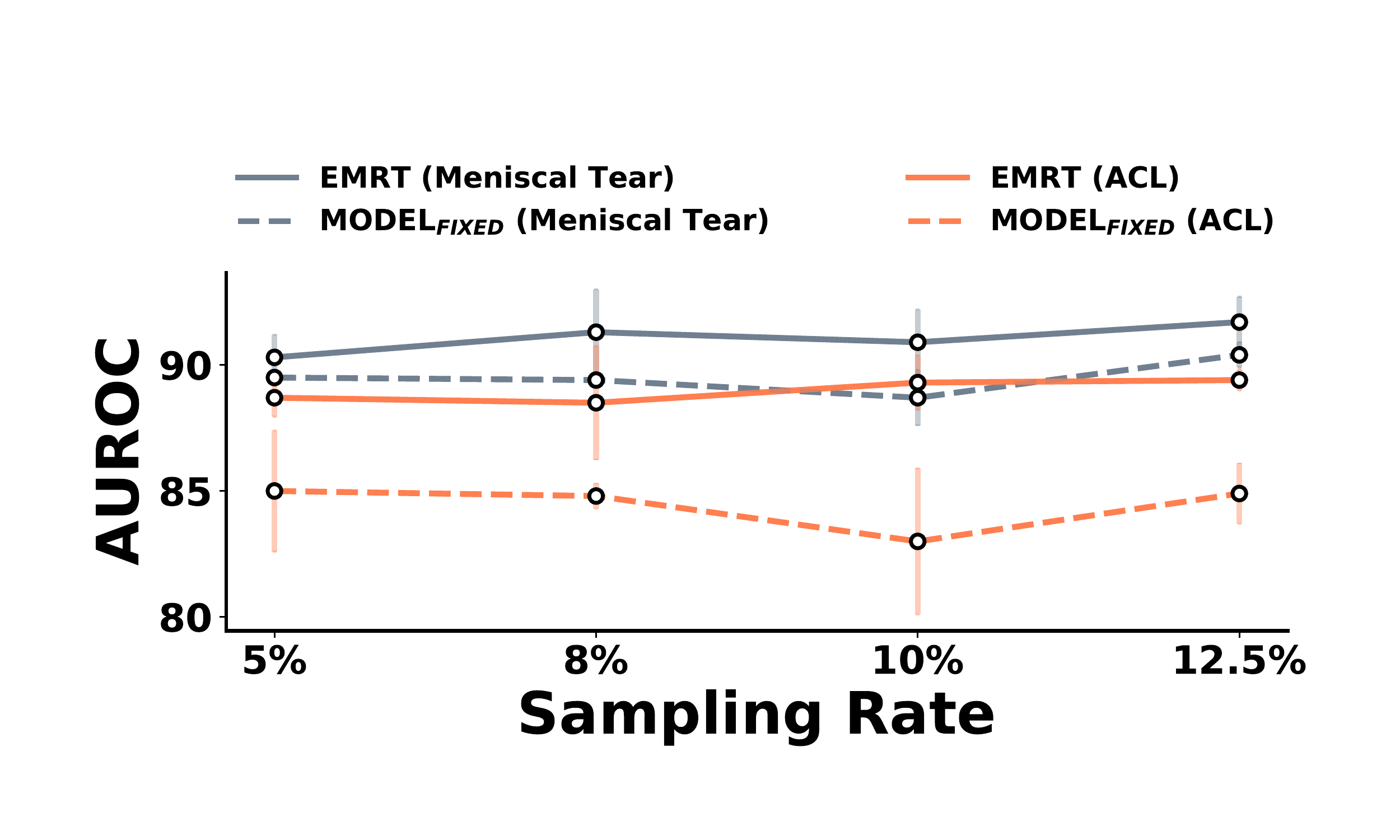}\includegraphics[width=0.50\textwidth]{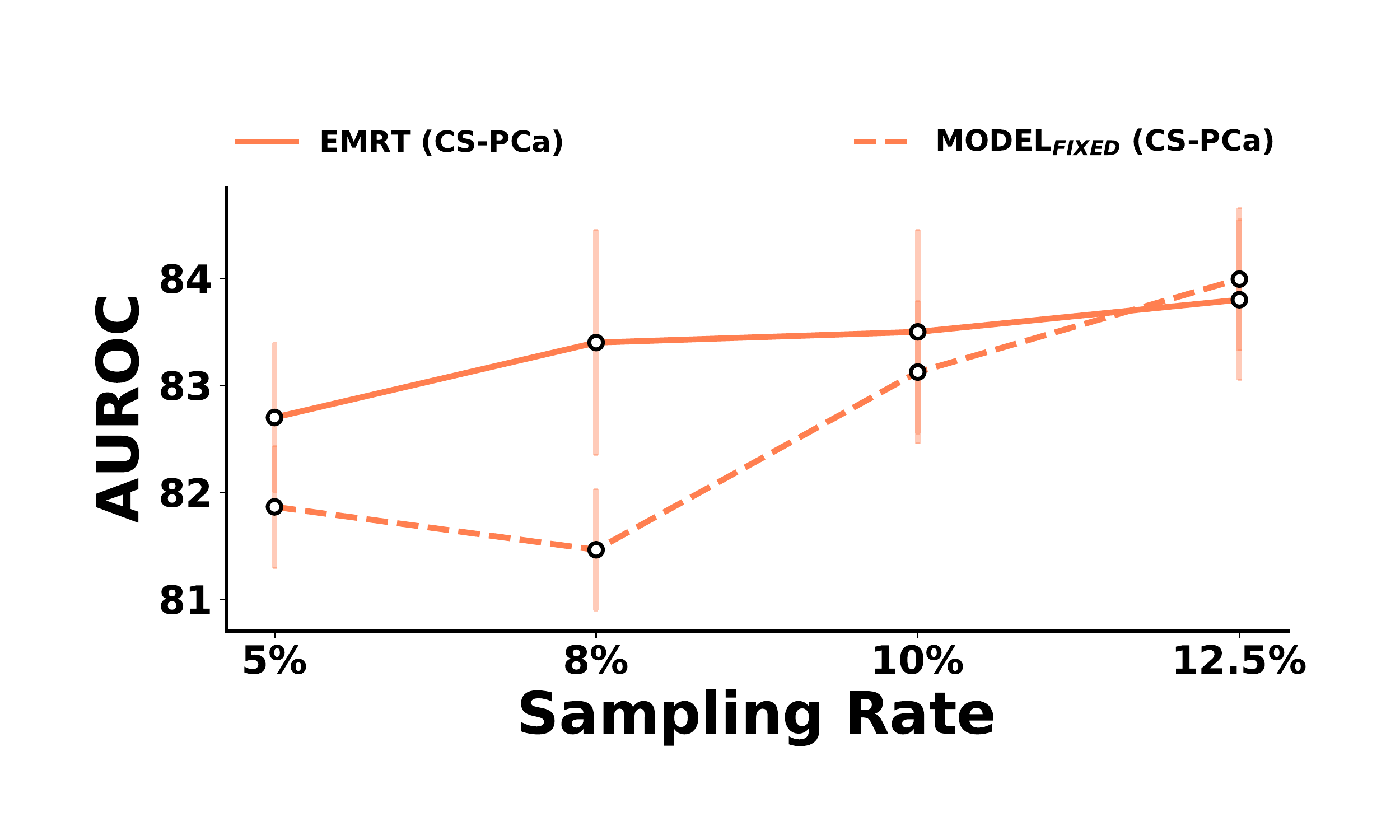}
\end{center}
\includegraphics[width=0.50\textwidth]{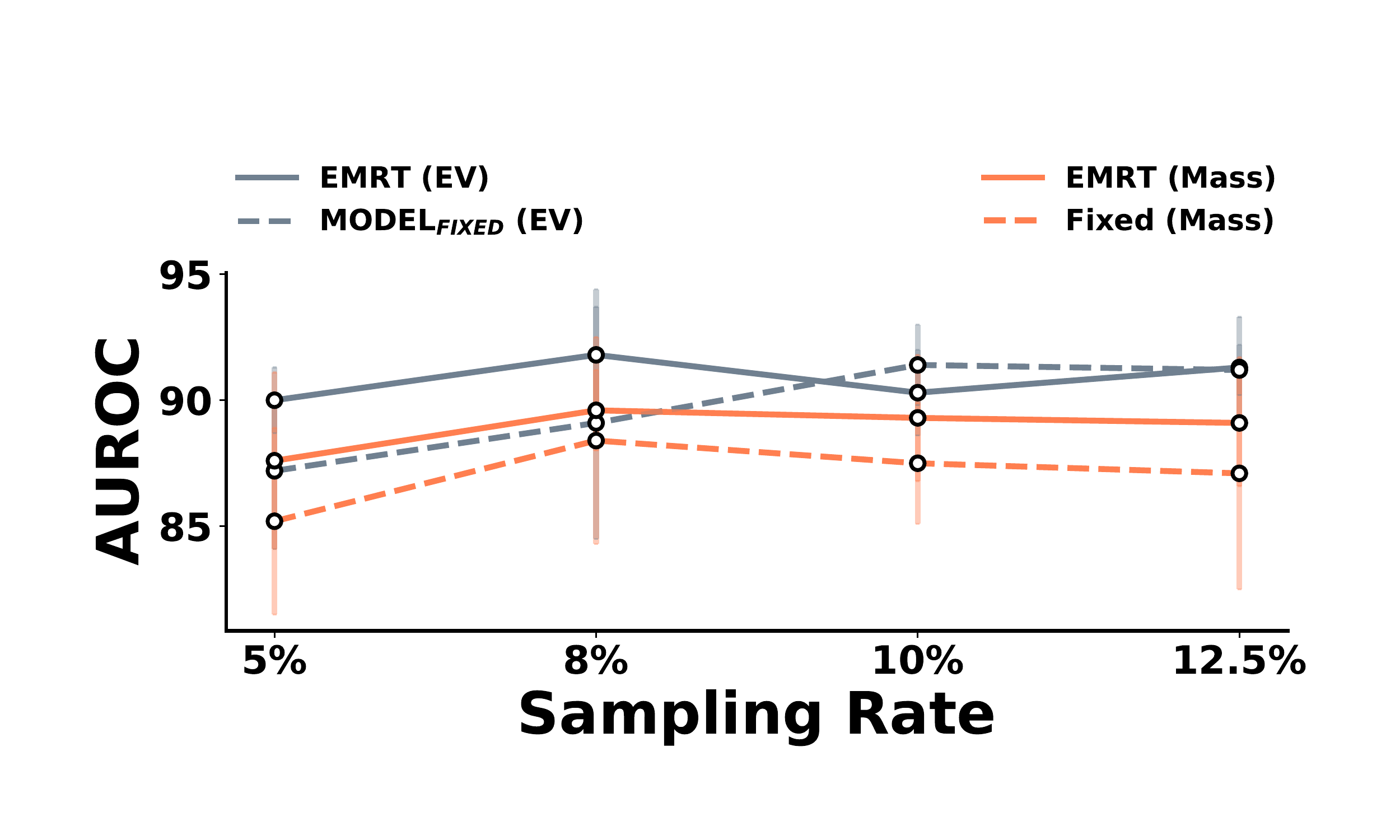}
\caption{\textit{\textbf{The Role of Random Subset Training in \textsc{emrt}}. Compares the classification 
performance of the \textsc{kspace-net} trained
using the \textsc{emrt} under-sampling patterns (dashed lines), \textsc{model}$_\textsc{fixed}$,
against \textsc{emrt} (solid lines).}}
\label{fig:empirical_1}
\end{figure}

One of the key characteristics of the training methodology of \textsc{emrt} is the way the \textsc{kspace-net} model $q_{\text{val}}$ is trained. 
Specifically, during the training of the classifier $q_{\text{val}}$, every mini-batch is constructed by first randomly drawing a different sampling pattern from the distribution $\pi$, and then applying the chosen pattern to all the samples in the mini-batch (see \Cref{alg:qval_training}). 
To better understand the role of this specialized training procedure on the performance of \textsc{emrt}, we examine whether training a \textsc{kspace-net} classifier using different sampling patterns across different mini-batches has any benefit compared to training a classifier trained using the same fixed sampling pattern across mini-batches. 
To that end, we compare the performance of the \textsc{emrt} classifier $q_{\text{val}}$ to a model trained with the fixed but learnt sampling pattern.
We use the sampling pattern learnt by \textsc{emrt} as the input to this classifier.
The architecture of the two classifiers were identical.
In \Cref{fig:empirical_1}, we observe that for most sampling rates the classifier trained using different sampling patterns across mini-batches outperforms the classifier trained with a single fixed sampling pattern, even if the fixed pattern is learnt. 
Training using the randomly chosen sampling patterns across mini-batches act as a regularizer which leads to better generalization performance.

\section{Conclusion and Limitations}
\textsc{mr} imaging is the gold standard of diagnostic imaging, especially in a differential diagnosis setting, thanks to its excellent soft-tissue contrast properties. 
However, despite its proven diagnostic value, this imaging modality is not used as a first-in-line tool for early identification of life threatening diseases, primarily because of lack of accessibility of this modality at population level. 
This lack of accessibility can be attributed to the need to generate high-fidelity images that are examined by radiologists. 
This is so because high-fidelity image generation necessitates the use of expensive scanning hardware to acquire large quantities of high quality \kspace data and the execution of complex and time consuming acquisition protocols to collect this data. 
Motivated by the goal of improving accessibility of \textsc{mr} for early and accurate disease identification at the population level,
in this study we propose to skip the image reconstruction step and instead propose to infer the final answer (presence/absence of the 
disease) directly from the \kspace data. 
We hypothesize that when image reconstruction is not a requirement, one can infer the presence/absence of the disease using a very 
small tailored fraction of the \kspace data. 
Towards that end we propose a novel deep neural network methodology, which we call \textsc{emrt}  that first learns the subset of the 
\kspace data which has the largest diagnostic signal to infer the disease and then uses this data to directly infer the disease 
without generating images. 
We validate our hypothesis by running a series of experiments using small sampling rates without suffereing a significant drop in performance compared to models using the fully-sampled \kspace. Models such as \textsc{emrt} that infer the presence of a disease directly from the \kspace data have the potential to bring \textsc{mr} scanners closer to deployment for population-level screening of disease.

\paragraph{Limitations} Despite encouraging preliminary results, much work needs to be done to get us closer to a system that can be clinically deployed. 
The present work is just a first step towards assessing the feasibility of whether it is possible to accurately infer the presence of the disease from a small tailored fraction of \kspace data without generating images.  
There are several limitations associated with the current work, which need to be  addressed to bring us closer to developing an actual scanning hardware that can operate outside of the specialized imaging environments and yet capture sufficient quantity and quality of the \kspace data for the subsequent \textsc{ml} model to infer the disease accurately. 

First, the current study works with the data generated from an expensive high-field $3$T scanner (the current standard of care) which is housed in specialized imaging environments. 
As a result the underlying \kspace data is of very high quality. 
In order for these results to generalize to the data acquired by more accessible low-field scanners, one needs to account for the noise ingrained in the data acquired by these low-field scanners. 
The current work does not propose any mechanism to account for such noise. 
It only focuses on establishing the limits on the quantity of data needed for accurate diagnosis. 

Second, almost all the modern day scanners acquire data in parallel using multiple coils.
This not only speeds up the data acquisition process but also increases the signal-to-noise (\textsc{snr}) ratio of the acquired signal.
However, in the current feasibility study, for the sake of simplicity, we resorted to working with the \textsc{esc} data (the multi-coil data emulated to be coming from a single coil). 
Future work will focus on extending the  \textsc{emrt} methodology for the multi-coil \kspace data.
We anticipate that working with multi-coil data will only lead to an improvement in performance because of the larger effective \textsc{snr} associated with the multi-coil data.  

Third, \textsc{mr} imaging is a $3$D imaging modality, where the human clinician renders the disease diagnosis after looking at all the slices in the volumetric image. 
The individual slices are seldom interpreted in isolation. 
In other words the final diagnosis is at the volume-level.  
However, in the current study, because of a dearth of positive cases at volume-level in our data set, we developed the \textsc{emrt} methodology to classify individual slices. 
Volume-level labels can be derived from labels of individual slices within the volume using any aggregation scheme, such as majority voting or averaging the probabilities of individual slices. 
However, naively aggregating slice-level labels can potentially lead to an increase in the number of false positive volumes. 
As part of the future work, with the help of additional data, we will explore extending the \textsc{emrt} methodology to directly classify the volumes. 

Another limitation of \textsc{emrt} comes from its use of the type of \kspace data. 
In a typical clinical \textsc{mr} scan multiple volumetric images are reconstructed, each having different contrast properties, with the goal of providing a radiologists with multiple visual facets of the same underlying anatomy. 
These different contrast images are reconstructed from the \kspace data corresponding to different acquisition sequences. 
For instance, prostate scans are typically acquired using T2-weighted (T2) and Diffusion-weighted (DW) sequences. 
However, again in the interest of simplicity, the \textsc{emrt} methodology proposed in this study uses the \kspace data from a single sequence. 
In the future we plan to extend this methodology to incorporate data from multiple sequences informed by what is used in real clinical settings. 
Lastly, the \textsc{emrt} methodology is restricted to learning only the Cartesian sampling patterns. 
However, for a given disease identification accuracy, there might exist other non-Cartesian sampling patterns which are  even sparser than the corresponding Cartesian pattern. 
While learning such ``arbitrary'' sampling patterns one needs to restrict to sample from the subset of patterns that respect the physical constraints of the scanner. 
In our future work we will also extend \textsc{emrt} to learn such ``arbitrary'' sampling patterns. 
Furthermore, to facilitate further research in this potentially high impact area, we are releasing a repository containing the data set and code for reproducing the experiments.

\bibliography{kspace2answer_main}
\bibliographystyle{plain}

\appendix

\section{Classification Metrics}\label{appsec:metrics}
\subsection{Knee Results}

\begin{table}[H]
\centering
\begin{tabular}{@{}clcccc@{}}
\toprule
\textbf{Sampling Rate} & \textbf{Pathologies} & {NPV / PPV (ARMS)}  & NPV / PPV (Recon) &  NPV / PPV (RSS) \\ \midrule
{$100\%$}   & ACL           &    &  & 99.2 ± 0.2 / 14.5 ± 0.9 \\ \cmidrule(l){2-6} 
                       & Meniscal Tear &   &  & 97.4 ± 0.3 / 44.9 ± 4.7 &  \\ \midrule
{$12.5\%$}   & ACL           & 99.1 ± 0.1 / 13.1 ± 1.5  & 98.8 ± 0.3 / 10.9 ± 1.7  &  \\ \cmidrule(l){2-6} 
                       & Meniscal Tear & 97 ± 0.3 / 42.3 ± 1.7   & 96.9 ± 0.5 / 10.9 ± 1.7  &  \\ \midrule
{10\%} & ACL           & 99.3 ± 0.2 / 12.7 ± 1.3 & 98.9 ± 0.2 / 11.5 ± 2   &     \\ \cmidrule(l){2-6} 
                       & Meniscal Tear & 97.6 ± 0.5/ 41.1 ± 2 & 97.1 ± 0.6 / 33.9 ± 2.5 &   \\ \midrule
{8\%}   & ACL           & 99. ± 0.3 / 13 ± 1.2 & 99 ± 0.2 / 11.1 ± 1.6  &  \\ \cmidrule(l){2-6} 
                       & Meniscal Tear & 97.8 ± 0.4 / 41.1 ± 2 & 97.1 ± 0.4 / 33.9 ± 2.4  &  \\ \midrule
{5\%}    & ACL           & 99.1 ± 0.1 / 13.3 ± 0.7  & 98.8 ± 0.3 / 11.5 ± 1.6  &  \\ \cmidrule(l){2-6} 
                       & Meniscal Tear & 97. ± 0.3 / 39.8 ± 1.3  & 96.8 ± 0.5 / 34.4 ± 2.9 &  \\ \bottomrule
\end{tabular}
\caption{Knee NPV/PPV Results}
\end{table}

\begin{table}[H]
\centering
\begin{tabular}{@{}clcccc@{}}
\toprule
\textbf{Sampling Rate} & \textbf{Pathologies} & {Sens / Spec (ARMS)}  & Sens / Spec (Recon) &  Sens / Spec (RSS) \\ \midrule
{100\%}   & ACL           &    &  & 81.1± 4.4 / 82.2± 2.2 \\ \cmidrule(l){2-6} 
                       & Meniscal Tear &   &  & 82.8± 2.2 / 86± 2.7 &  \\ \midrule
{12.5\%}    & ACL    & 80.9 ± 4.6 / 79.7 ± 4.2 & 75.5 ± 8.2 / 76.2 ± 7.3 &  \\ \cmidrule(l){2-6} 
                       & Meniscal Tear & 82.2 ± 2.5 / 84.8 ± 0.8 & 81.4 ± 3.1 / 78 ± 2.8 &  \\ \bottomrule
{10\%}   & ACL      & 80 ± 2.6  / 80.7 ± 1.1 & 77 ± 4.1  / 77.2 ± 4.9 &  \\ \cmidrule(l){2-6} 
                       & Meniscal Tear & 81 ± 1.9  / 83.4 ± 1.2 & 82.8 ± 3.7 / 78 ± 2.3  &  \\ \midrule
{8\%} & ACL    & 78.2 ± 7.3 / 80.5 ± 2 & 80.2 ± 4.2 / 75.6 ± 4.6  &     \\ \cmidrule(l){2-6} 
                       & Meniscal Tear & 80.6 ± 2.2 / 84.4 ± 0.9 & 82.8 ± 2.8 / 78.1 ± 2.4 &   \\ \midrule
{5\%}   & ACL           & 84.8 ± 3.5 / 78.1 ± 2.5 & 73.9 ± 8.2 / 78.1 ± 6.3 &  \\ \cmidrule(l){2-6} 
                       & Meniscal Tear & 80.8 ± 3.4 / 84 ± 1.1 & 81 ± 3.2  / 78.9 ± 2.8  &  \\ \midrule
\end{tabular}
\caption{Knee Sensitivity / Specificity Results}
\end{table}

\subsection{Brain Results}

\begin{table}[H]
\centering
\begin{tabular}{@{}clcccc@{}}
\toprule
\textbf{Sampling Rate} & \textbf{Pathologies} & {NPV / PPV (ARMS)}  & NPV / PPV (Recon) &  NPV / PPV (RSS) \\ \midrule
{100\%}   & Enlarged Ventricles           &    &  & 99.6 ± 0.2 / 18.3 ± 9.7 \\ \cmidrule(l){2-6} 
                       & Mass &   &  & 99.5± 0.3 / 8.1± 0.9 &  \\ \midrule
{12.5\%}   & Enlarged Ventricles    & 99.5 ± 0.1 / {15.3 ± 7.1}  & 99.3 ± 0.3 / 5.9 ± 1.5  &  \\ \cmidrule(l){2-6} 
                       & Mass & 99.5 ± 0.2 / {8.3 ± 1.4}  & 98.8 ± 0.2/  3.8  &  \\ \midrule
{10\%} & Enlarged Ventricles           & 99.5 ± 0.1 / 11.3 ± 4 & 99.4 ± 0.1/ 8.1 ± 2.5  &     \\ \cmidrule(l){2-6} 
                       & Mass & 99.4 ± 0.2 / 6.7 ± 1.4  & 99.4 ± 0.2/ 5.1 ± 1.1 &   \\ \midrule
{8\%}   & Enlarged Ventricles           & 99.6 ± 0.1 / 9.3 ± 3.7 & 99.4 ± 0.2/ 5.1 ± 1.1  &  \\ \cmidrule(l){2-6} 
                       & Mass & 99.6 ± 0.1 / 6.8 ± 1.3 & 98.7 ± 0.2/ 4.4 ± 0.8  &  \\ \midrule
{5\%}    & Enlarged Ventricles           & 99.5 ± 0.1 / 9.1 ± 2.2  & 99.4 ± 0.2/ 6.5 ± 2.2  &  \\ \cmidrule(l){2-6} 
                       & Mass & 99.5 ± 0.3 / 7 ± 1.2  & 98.7 ± 0.2 / 4.4 ± 0.6 &  \\ \bottomrule
\end{tabular}
\caption{Brain NPV/PPV Results}
\end{table}

\begin{table}[H]
\centering
\begin{tabular}{@{}clcccc@{}}
\toprule
\textbf{Sampling Rate} & \textbf{Pathologies} & {Sens / Spec (ARMS)}  & Sens / Spec (Recon) &  Sens / Spec (RSS) \\ \midrule
{100\%}   & Enlarged Ventricles           &    &  & 84.9 ± 7 / 85.8 ± 7.9 \\ \cmidrule(l){2-6} 
                       & Mass &   &  & 86.2 ± 6.9 / 72.4 ± 3.8 &  \\ \midrule
{12.5\%}    & Enlarged Ventricles    & 83.3 ± 2.2 / 84.3 ± 7.5 & 83.4 ± 8.1 / 62.2 ± 11.7  &  \\ \cmidrule(l){2-6} 
                       & Mass & 85.6 ± 4.7 / 73 ± 4.8 & 83.3 ± 5   / 38.8 ± 12.2 &  \\ \bottomrule
{10\%}   & Enlarged Ventricles      & 83.9 ± 4.1 / 79 ± 10.4 & 84.1 ± 5   / 71.9 ± 10.3 &  \\ \cmidrule(l){2-6} 
                       & Mass & 85.9 ± 4.9 / 65.3 ± 7.8 & 73.9 ± 4.6 / 56.3 ± 7.7 &  \\ \midrule
{8\%} & Enlarged Ventricles    & {88.2 ± 3.7} / 74.1 ± 8.2  & 88.5 ± 3.7 / 54.2 ± 11.1  &     \\ \cmidrule(l){2-6} 
                       & Mass & {90.0 ± 2.1}/64.7 ± 5.5 & 74.2 ± 5.4 / 53.5 ± 11.1 &   \\ \midrule
{5\%}   & Enlarged Ventricles           & {86.2 ± 4.5}/75.4 ± 7.4 & 84.8 ± 7.9 / 63.1 ± 14.8 &  \\ \cmidrule(l){2-6} 
                       & Mass & {87.8 ± 7.7}/66.3 ± 7.2 & 73.4 ± 3.9 / 55.2 ± 7.5  &  \\ \midrule
\end{tabular}
\caption{Brain Sensitivity / Specificity Results}
\end{table}

\subsection{Prostate Results}

\begin{table}[H]
\centering
\begin{tabular}{@{}clcccc@{}}
\toprule
\textbf{Sampling Rate} & \textbf{Pathologies} & {Sens / Spec (ARMS)}  & Sens / Spec (Recon) &  Sens / Spec (RSS) \\ \midrule
{100\%}   & CS-PCa           &    &  & 93.3 ± 0.5 / 59.3 ± 5.4 \\ 
     \midrule
{12.5\%}   & CS-PCa  & 91.1 ± 9.6 / 59.2 ± 1.9  & 90 ± 9.6   / 57.9 ± 1.9  &  \\\midrule
{10\%} & CS-PCa     & 88 ± 8.1   / 64.7 ± 5.1 & 86 ± 8.1   / 54.4 ± 2.3 &   \\ \midrule
{8\%}   & CS-PCa   & 91.3 ± 5.3 / 60.8 ± 2.1 & 89 ± 5.3   / 54.3 ± 2.1   \\ \midrule
{5\%}    & CS-PCa  & 88.5 ± 4.4 / 62.9 ± 1.5  & 88.6 ± 4.4 / 47 ± 1.5  \\ \bottomrule
\end{tabular}
\caption{Prostate Sensitivity/Specificity Results}
\end{table}

\begin{table}[H]
\centering
\begin{tabular}{@{}clcccc@{}}
\toprule
\textbf{Sampling Rate} & \textbf{Pathologies} & {NPV / PPV (ARMS)}  & NPV / PPV (Recon) &  NPV / PPV (RSS) \\ \midrule
{100\%}   & CS-PCa           &    &  & 99.2 ± 0.0 / 14.5 ± 1.6 \\ 
     \midrule
{12.5\%}   & CS-PCa  & 98.7 ± 0.2 /13.4 ± 1.8  & 98.7 ± 0.6 / 12.2 ± 1.8   &  \\\midrule
{10\%} & CS-PCa     & 99 ± 0.3 /12.8 ± 5 & 98.8 ± 0.6 / 11.7 ± 5   &   \\ \midrule
{8\%}   & CS-PCa   & 98.7 ± 0.6  /  13.8 ± 2.1 & 97 ± 0.3 /11.8 ± 2.1  \\ \midrule
{5\%}    & CS-PCa  & 98.9 ± 0.6 / 12.1 ± 1.5 & 96.9 ± 0.1 /10 ± 1.5   \\ \bottomrule
\end{tabular}
\caption{Prostate NPV/PPV Results}
\end{table}

\end{document}